%% file: braggAI.tex
\documentclass[final]{amsart}
\usepackage{Aaron_Pim_Style}
\usepackage{booktabs, fullpage, algorithm, algpseudocode, pgfplots, pgfplotstable, subcaption}
\usepackage{tikz}
\usetikzlibrary{positioning, arrows.meta, calc,snakes}
\usepackage[giveninits=true,eprint=false,url=false,style=alphabetic]{biblatex}

\usepackage{caption}
\newcommand{\aaron}[1]{\textcolor{blue}{Aaron says: #1}}

\usepackage[dvipsnames]{xcolor}
\usepackage[colorlinks=true, linkcolor=Purple, citecolor=Orange, urlcolor=Purple]{hyperref}

\title{Surrogate Modelling of Proton Dose with Monte Carlo Dropout
  Uncertainty Quantification}

\author[2]{Aaron Pim}

\author[1,2]{Tristan Pryer}

\address{$^1$ Institute for Mathematical Innovation\\ University of
  Bath, Bath, UK. $^2$ Department of Mathematical Sciences
  \\ University of Bath, Bath, UK.}

\begin{document}

\maketitle

\begin{abstract}
  Accurate proton dose calculation with Monte Carlo (MC) remains
  computationally demanding in workflows that require repeated
  evaluations, such as robust optimisation, adaptive replanning and
  probabilistic inference. We construct a neural surrogate that
  incorporates Monte Carlo dropout to provide fast, differentiable
  dose predictions together with voxelwise predictive uncertainty. The
  method is validated in a staged series of experiments, a
  one-dimensional analytic benchmark establishes accuracy, convergence
  and variance decomposition; two-dimensional bone-water phantoms
  generated with TOPAS/Geant4 demonstrate behaviour under domain
  heterogeneity and beam uncertainty and a three-dimensional water
  phantom confirms scalability to volumetric dose prediction. Across
  settings we separate epistemic (model) from parametric (input)
  contributions, showing that epistemic variance inflates under
  distribution shift while parametric variance dominates at material
  boundaries. 
  The approach achieves orders-of-magnitude speedup over
  MC while retaining uncertainty information and is intended for
  integration into robust planning, adaptive workflows and
  uncertainty-aware optimisation in proton therapy.
\end{abstract}

\section{Introduction}

Proton beams deposit most of their energy near the end of range,
producing a distal Bragg peak; small changes in tissue composition or
density shift this peak and alter dose to targets and organs at risk,
so accurate dose calculation is central to planning
\cite{liu2011proton}. Deterministic formulations, based on the
continuous slowing down approximation or transport equations, capture
average behaviour and admit analytic approximations, but they neglect
statistical fluctuations
\cite{burlacu2023deterministic,ashby2025positivity}. Stochastic
formulations, by contrast, describe individual particle paths,
accounting for deterministic energy loss from inelastic interactions
and random angular deflections from Coulomb scatter, with additional
variability introduced by range straggling
\cite{crossley2024jump,chronholm2025geometry,kyprianou2025unified}. Analytic
models of both types provide reduced-order descriptions of dose, while
numerical approaches range from PDE solvers and pencil-beam algorithms
to Monte Carlo and SDE-based simulations
\cite{ashby2025efficient,NewhauserZhang:2015}. Monte Carlo (MC)
transport remains the reference standard for accuracy but is
computationally demanding even when GPU implementations and clinical
verifiers are used \cite{giantsoudi2015validation,zhou2024clinical}.
To reduce wall time while retaining MC fidelity, recent work trains
deep surrogates to predict LET and dose with millisecond–second
runtimes, e.g. LET calculators and 3-D dose/LETD predictors trained on
MC or hybrid data
\cite{tang2024deep,pirlepesov2022three,pastor2022millisecond,starke2024deep}
and extends to heavy-ion therapy for online adaptation and rapid QA
\cite{he2025deep,he2025deep2}. Complementary denoisers map
low-particle MC to high-quality dose, shrinking simulation budgets
\cite{zhang2023deep}, and fast conversions lift pencil-beam solutions
to MC-quality dose in seconds for clinical use \cite{wu2021improving}.

These accelerations address speed, not trust. Deterministic predictors
provide point estimates only; uncertainty quantification (UQ) is
needed to audit reliability, enable robust optimisation
\cite{georgiou2025scotty}, dose delivery inference
\cite{cox2024bayesian} and guide data acquisition
\cite{wildman2025recent,staahl2020evaluation}. A practical route is
Monte Carlo dropout (MC-dropout). Dropout was introduced as a
regulariser that randomly masks activations during training to reduce
co-adaptation \cite{srivastava2014dropout} and later reinterpreted as
approximate Bayesian inference, so repeated stochastic test-time
passes yield predictive means and variances with minimal code changes
\cite{gal2016dropout}. Variants improve calibration and sample
efficiency \cite{hasan2023controlled} and applications in medical
imaging show that uncertainty highlights failure modes and supports
downstream decisions
\cite{sahlsten2024application,klanecek2023uncertainty}. These
ingredients motivate an uncertainty-aware surrogate pipeline that
scales from one-dimensional depth-dose to full three-dimensional dose.

Clinically, the appeal of protons is precisely the steep distal
fall-off around the Bragg peak. That strength is also a vulnerability,
millimetric errors in water-equivalent path length, unmodelled
heterogeneity, or small setup shifts can displace the high-dose region
relative to target and organs at risk. In practice this means that
accuracy at the distal edge is not only a numerical goal but a
planning requirement, since misplacement of the peak risks target
under-dosage or excess dose to critical structures. This sensitivity
concentrates the need for trustworthy predictions where gradients are
largest and where tissue changes most strongly affect range
\cite{liu2011proton}.

Despite advances in GPU implementations and clinically validated
verifiers \cite{giantsoudi2015validation,zhou2024clinical}, full MC
remains costly for workflows that require many evaluations. Modern
planning iterates dose engines thousands of times, robust optimisation
evaluates scenarios across range and setup perturbations, adaptive
workflows revisit dose after anatomical change and probabilistic
inversion or Bayesian calibration loops demand repeated forward
solves. Denoisers \cite{zhang2023deep} and fast conversions from
analytic models \cite{wu2021improving} mitigate per-evaluation cost,
yet the cumulative budget for high-fidelity MC still restricts the
breadth of scenario sets and the use of sampling-based UQ within tight
clinical time frames.

Fast surrogates promise a complementary path. By learning the map from
beam and medium parameters to dose, a differentiable emulator can be
embedded in inner optimisation loops, enable sensitivity analysis with
automatic differentiation and support sampling-based analyses at
interactive speeds. This aligns with emerging applications in online
adaptation, rapid QA and heavy-ion settings
\cite{tang2024deep,pirlepesov2022three,pastor2022millisecond,starke2024deep,he2025deep,he2025deep2}. However,
point predictions alone are insufficient for safe decision
making. Robust planning, data-efficient acquisition and model auditing
all require uncertainty estimates that are well calibrated and that
respond sensibly to distribution shift
\cite{wildman2025recent,staahl2020evaluation}.

This creates an unmet need for uncertainty-aware fast dose predictors,
these models approach MC fidelity in nominal cases, expose voxelwise
uncertainty that inflates at distal fall-off and material interfaces,
and remain simple enough to deploy within existing planning
stacks. MC-dropout offers a pragmatic solution
\cite{gal2016dropout,srivastava2014dropout}. It preserves the usual
training and inference toolchain, yields test-time ensembles with
minimal code changes and retains compatibility with automatic
differentiation for optimisation. Controlled variants can improve
calibration \cite{hasan2023controlled} and prior literature in medical
imaging suggests that the resulting uncertainty maps can flag likely
failure modes and guide downstream choices
\cite{sahlsten2024application,klanecek2023uncertainty}. In this work
we adopt MC-dropout to construct a surrogate pipeline that runs from
1-D depth-dose to 3-D dose, quantifies both model and input
variability and incorporates simple post-hoc calibration so nominal
and empirical coverages agree.

\subsection{Contribution of the work}

We construct a neural surrogate for proton dose that exposes
calibrated predictive uncertainty through MC-dropout while retaining
automatic differentiation for optimisation and inference. Our approach
proceeds in stages of increasing dimensionality and complexity.

We begin with a one-dimensional analytic benchmark of depth-dose
profiles using the model from \cite{ashby2025efficient}. This
controlled setting allows us to test the surrogate against a model
with closed-form behaviour, establish the accuracy of mean
predictions, and examine sharpness and empirical coverage of credible
intervals. Mathematically, the 1-D case provides a clean setting for
variance decomposition and convergence studies (in training samples
and dropout passes), while clinically it corresponds to the core
range-dose trade-off at the Bragg peak that underlies proton therapy
\cite{liu2011proton}.

We then extend to two-dimensional log-projection maps in a controlled
bone-water phantom. Here, the surrogate is trained on MC-generated
data from TOPAS/Geant4, with uncertainty in bone position and
thickness capturing heterogeneity effects. This stage demonstrates
that the surrogate generalises from analytic inputs to realistic
voxelised MC data, that variance concentrates at material boundaries
and distal fall-off, and that epistemic and parametric components can
be disentangled. Clinically, it mimics common scenarios where
interfaces (e.g. bone-soft tissue) perturb range and motivate robust
margins.

Finally, we move to three-dimensional voxel dose in a homogeneous
water phantom with perturbed beam setup. This tests scalability of the
method to full volumetric data and shows that uncertainty quantification
remains tractable at clinical resolutions. It highlights how epistemic
uncertainty localises at the distal Bragg surface while parametric
uncertainty reflects beam configuration variability. For clinical
practice this demonstrates feasibility of embedding an uncertainty-aware
surrogate within adaptive or robust planning pipelines.

Across all stages we quantify and disentangle epistemic uncertainty
from parametric input variability, validate behaviour under distribution
shift, and apply simple post-hoc calibration so nominal and empirical
coverages agree. The approach integrates naturally with accelerated and
denoised MC pipelines
\cite{giantsoudi2015validation,zhou2024clinical,zhang2023deep} and with
learned dose and LET surrogates
\cite{tang2024deep,pirlepesov2022three,pastor2022millisecond,starke2024deep,wu2021improving},
providing a coherent framework for robust planning and UQ.

\subsection{Relation to the literature}

The work sits at the intersection of fast yet accurate dose
computation and practical UQ for deep surrogates. On the computation
side, GPU MC and clinically deployed verifiers reduce wall time but
still incur costs that scale with repeated evaluations. Deep
surrogates and denoisers compress runtimes by orders of magnitude with
high gamma pass rates, and have been demonstrated for proton and heavy
ion dose/LETD prediction, denoising and fast pencil beam corrections
\cite{giantsoudi2015validation,zhou2024clinical,zhang2023deep,pastor2022millisecond,starke2024deep,wu2021improving,tang2024deep,he2025deep,he2025deep2}. Newer
work extends these ideas to Bayesian networks and synthetic CT
pipelines. BayesDose uses Bayesian LSTMs with weights drawn from
Gaussian mixture models to produce ensemble predictions, showing that
100 ensemble passes yield mean predictions comparable to deterministic
LSTMs and that the resulting predictive standard deviation correlates
with dosimetric errors while the runtime overhead can be reduced to
$\approx9\times$ that of a single forward pass
\cite{voss2023bayesdose}. In adaptive workflows, Monte Carlo dropout
based uncertainty maps on deep learning synthetic CTs correlate
strongly with HU, range, WET and dose errors, demonstrating the
utility of uncertainty maps as QA tools for online adaptive proton
therapy \cite{galapon2024feasibility}. Complementary approaches
directly estimate uncertainty by reconstructing the input \cite{huet2024can}.

On the UQ side, evaluation frameworks for deep learning highlight the
importance of calibration and coverage guarantees
\cite{staahl2020evaluation}, and Bayesian segmentation and MC dropout
studies in oncology show that test‑time sampling captures epistemic
effects that correlate with error and can be used to screen
predictions
\cite{sahlsten2024application,klanecek2023uncertainty}. Bayesian
neural networks and ensemble methods provide alternative UQ
approaches; for example, BayesDose samples network weights from
learned distributions to estimate mean dose and variance
\cite{voss2023bayesdose}, while direct reconstruction methods estimate
uncertainty without multiple stochastic passes
\cite{huet2024can}. Calibrated conformal methods and controlled
dropout variants can further improve coverage and reliability
\cite{hasan2023controlled}. Reviews of machine learning for proton
radiotherapy emphasise both the opportunity and the need for
principled UQ in model based pipelines \cite{wildman2025recent}. We
adopt MC dropout for its simplicity and scalability
\cite{srivastava2014dropout,gal2016dropout} and note that more
sophisticated Bayesian or reconstruction based methods could be
substituted in future work. Together these strands motivate and inform
the uncertainty aware surrogate design presented here, which seeks to
bridge fast dose computation with reliable, calibrated uncertainty
estimates.

The rest of the paper is organised as follows.
Section~\ref{sec:background} summarises the relevant proton beam
physics, dose calculation by Monte Carlo, and the motivation for
uncertainty-aware surrogates.  Section~\ref{sec:problemsetup} sets out
the surrogate formulation, including network architecture, Monte Carlo
dropout, variance decomposition, and calibration methodology.
Section~\ref{sec:numerics} presents numerical experiments, beginning
with foundational one-dimensional analytic benchmarks and progressing
to higher-dimensional phantom studies.  Section~\ref{sec:discussion}
discusses the results in both mathematical and clinical terms,
emphasising sources of uncertainty, computational trade-offs, and
behaviour under distribution shift.  Finally,
Section~\ref{sec:outlook} summarises the main findings, notes current
limitations, and outlines directions for future work.

\section{Background physics and computational model}
\label{sec:background}

\subsection{Proton beam physics in brief}

Proton beams deposit energy primarily through inelastic interactions
with electrons. The macroscopic rate of energy loss along track length
$\ell$ is governed by the stopping power $S(E)$ via
\begin{equation}
-\frac{\mathrm d E}{\mathrm d \ell}=S(E),
\end{equation}
which increases as energy decreases, producing a pronounced distal
Bragg peak in depth–dose. Small changes in material composition and
density alter the water equivalent path length, shifting the peak and
amplifying sensitivity to heterogeneity. In practical therapy energies
the transport domain excludes $E\to 0$ where $S(E)$ becomes singular;
we work on $E\in[E_{\min},E_{\max}]$ with $E_{\min}>0$. Forward-peaked
multiple scattering contributes lateral spread that grows with depth
and depends on material, further coupling geometry and dose placement
\cite{liu2011proton}.

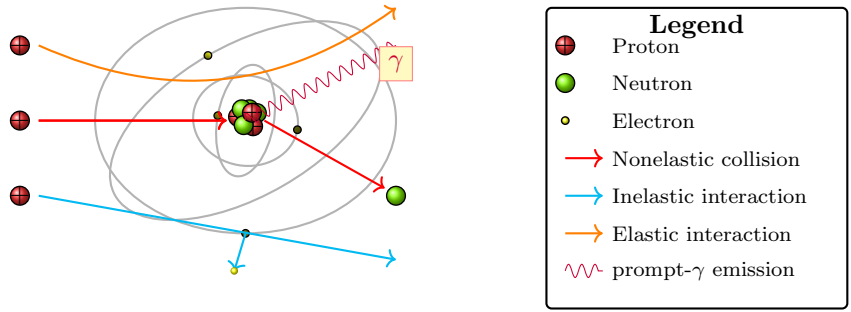
\begin{figure}[h!]
    \centering
    \input{nuclear.tex}
    \caption{\em The main interaction channels of a proton with
    matter: \textcolor{red}{nonelastic} proton–nucleus collisions,
    \textcolor{cyan!80!white}{inelastic} Coulomb interactions with
    atomic electrons, and \textcolor{orange}{elastic} Coulomb
    scattering with nuclei. \label{fig:atom}}
\end{figure}

Within the clinical energy range (50-150 MeV), the dominant
interactions are illustrated in Figure~\ref{fig:atom}. Inelastic
collisions with electrons cause gradual energy loss, typically
described deterministically through the Bragg-Kleeman or Bethe-Bloch
equations and give rise to the characteristic Bragg peak. Because
these collisions are discrete events, protons of identical initial
energy do not all stop at the same depth; this leads to longitudinal
spread of the peak, known as range straggling
\cite{bortfeld1997analytical}. Angular deflections occur primarily
through elastic Coulomb scattering with nuclei. These small but
frequent interactions accumulate to produce lateral beam broadening
via multiple scattering \cite{Gottschalk1993}. Less frequent inelastic
nuclear reactions generate secondary particles, notably neutrons, and
contribute to the distal halo of the dose distribution
\cite{Schneider2002}.

\subsection{Dose calculation and Monte Carlo}

From the physical processes described above, the key clinical
observable is the dose distribution, i.e. the spatial map of energy
deposition in the medium. For a given treatment configuration let
$\vec x$ collect the beam and medium parameters. The resulting dose
distribution on a fixed grid of voxels is denoted
$d(\vec x)\in\mathbb R^{M_1\times M_2\times \ldots}$, where each entry represents
the energy deposited per unit mass in the corresponding voxel.

At the particle level, Monte Carlo (MC) transport simulates individual
histories $Y^{(n)}$, $n=1,\ldots,N$, each of which is a stochastic
trajectory describing successive interactions of a proton with the
medium. Along a given history, let $\Delta E^{(n)}_k$ denote the
energy lost in the $k$th interaction, at spatial position
$X^{(n)}_k$. The indicator
$\chi_{\text{voxel}}(X^{(n)}_k)$ assigns this deposition to the voxel
that contains $X^{(n)}_k$. The exact dose can then be expressed as the
expectation
\begin{equation}
  d(\vec x)=\mathbb E \left[\sum_{k} \Delta E^{(n)}_k\,
  \chi_{\text{voxel}}(X^{(n)}_k)\right],
\end{equation}
where the sum runs over all interactions in a single history. In
practice this expectation is approximated by the sample mean over the
$N$ simulated histories,
\begin{equation}
  \label{eq:mc-estimator}
  d(\vec x) \approx \frac{1}{N}\sum_{n=1}^N
  \sum_{k} \Delta E^{(n)}_k \,\chi_{\text{voxel}}(X^{(n)}_k).
\end{equation}
The estimator variance scales like $\mathcal O(1/N)$, but each history
resolves many microscopic interactions and boundary crossings, so
wall-time is substantial even with GPU acceleration.

Deterministic approaches, such as pencil-beam algorithms or numerical
solvers for transport equations, are considerably faster, but they
rely on approximations that neglect heterogeneity effects or straggle
at the distal fall-off. These methods can provide useful first
estimates but may lack the fidelity required for high-precision
planning. MC therefore remains the reference standard for accuracy,
while its computational burden motivates the search for learned
surrogates that retain MC-like behaviour at inference speed
\cite{giantsoudi2015validation,pastor2022millisecond,staahl2020evaluation}.

\section{Problem Setup and Methodology}
\label{sec:problemsetup}

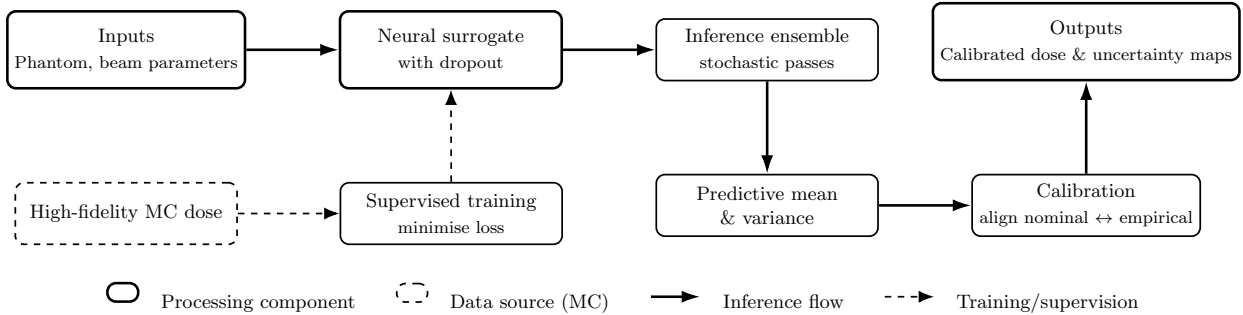
\begin{figure}[t]
\centering
\resizebox{\textwidth}{!}{\input{pipeline.tex}}
\caption{
  \label{fig:pipeline}
  Pipeline overview: inputs (phantom, beam parameters) $\to$
  neural surrogate with dropout $\to$ ensemble of stochastic passes
  $\to$ predictive mean and variance $\to$ calibration $\to$
  calibrated dose and uncertainty maps. A supervised training lane
  ingests Monte Carlo dose to fit the surrogate.}
\end{figure}

Before formalising the mathematics, we outline the pipeline in plain
terms. The inputs are phantom and beam parameters, e.g., tissue
composition, density, beam energy and angle. These parameters are fed
into a neural surrogate with dropout layers, trained on high-fidelity
data. At inference, repeated stochastic forward passes through the
surrogate yield not just a single dose prediction but an ensemble,
from which we compute a predictive mean and variance. A final
calibration step aligns the nominal confidence levels of these
uncertainty estimates with empirical coverage, ensuring that the
reported intervals are statistically reliable, as shown in
Figure~\ref{fig:pipeline}.

We now formalise the components shown in Figure~\ref{fig:pipeline}.
Let $\vec x\in\mathbb{R}^d$ collect beam and medium parameters
(energy, entry position, angle, material properties). Let
$\mathcal{Z}=\{z_j\}_{j=1}^{M}$ denote a fixed set of sampling
locations (depths in 1D, pixels in 2D, voxels in 3D). For a given
configuration $\vec x$, the reference dose is the discrete field
$\vec d(\vec x)\in\mathbb{R}^{M}$ defined on $\mathcal{Z}$, generated
by a high-fidelity simulation.

The surrogate is a parametric map
$\mathscr{D}_\theta:\mathbb{R}^d\to\mathbb{R}^{M}$ trained on a finite
dataset $\mathrm{D}_{\mathrm{train}}=\{(\vec x^{(i)}, \vec d(\vec
x^{(i)}))\}_{i=1}^{N}$.  Unless otherwise stated we use a quadratic
loss
\begin{equation}
  \mathcal{L}(\theta) = \frac{1}{N}\sum_{i=1}^N
  \|\mathscr{D}_\theta(\vec x^{(i)})-\vec d(\vec x^{(i)})\|_2^2
\end{equation}
with standard regularisation. At inference, dropout layers remain
active and a single stochastic forward pass yields
$\mathscr{D}^{(t)}_\theta(\vec x)$ for $t=1,\ldots,T$. The ensemble
mean and (epistemic) variance are estimated componentwise by
\begin{equation}
  \widehat\mu_j(\vec x)=\frac{1}{T}\sum_{t=1}^T \mathscr{D}^{(t)}_{\theta,j}(\vec x),
  \qquad
  \widehat\sigma^2_{\text{epi},j}(\vec x)=\frac{1}{T-1}\sum_{t=1}^T
   \bigl(\mathscr{D}^{(t)}_{\theta,j}(\vec x)-\widehat\mu_j(\vec x)\bigr)^2,
  \quad j=1,\ldots,M.
\label{eq:dropout-estimators}
\end{equation}
When input parameters are treated as uncertain, we model them as a
random vector $\vec X \sim \Pi$ with distribution $\Pi$ (for example,
capturing variability in material properties or beam configuration).
Independent samples $\vec x^{(s)} \sim \Pi$, $s=1,\ldots,S$, are then
drawn to form the nested estimator
\begin{equation}
  \widehat\mu_j=\frac{1}{S}\sum_{s=1}^S \widehat\mu_j(\vec x^{(s)}),\qquad
  \widehat\sigma^2_{\text{tot},j}=\underbrace{\frac{1}{S}\sum_{s=1}^S \widehat\sigma^2_{\text{epi},j}(\vec x^{(s)})}_{\text{epistemic}}
  + \underbrace{\frac{1}{S-1}\sum_{s=1}^S\bigl(\widehat\mu_j(\vec x^{(s)})-\widehat\mu_j\bigr)^2}_{\text{parametric}},
\end{equation}
realises the law of total variance at the discrete level and gives the
voxelwise decomposition reported in the experiments. Finally, a
split-conformal step rescales the half-widths of prediction intervals
so that nominal and empirical coverage agree on a held-out calibration
set.

\subsection{Network architecture}

The surrogate $\mathscr{D}_\theta$ is a feedforward neural network
mapping $\vec x \in \mathbb{R}^d$ to $\mathbb{R}^{M_1 \times M_2
  \times \ldots}$.  Its layers consist of an input transformation
$\mathcal{C}_{\mathrm{in}}$, a stack of hidden layers with ReLU
activation, optional dropout layers $\mathcal{C}_{d,i}$ in which
activations are multiplied by a Bernoulli mask
\begin{equation}
    \vec{B}_{p_\mathrm{drop}} \sim \text{diag}(\text{Bernoulli}(~\underbrace{(p_\mathrm{drop}, \ldots, p_\mathrm{drop})}_{N_{\text{width}}}~)).
\end{equation}
with retention probability $1-p_{\mathrm{drop}}$, and an output layer
$\mathcal{C}_{\mathrm{out}}$
\begin{equation}
\mathscr{D}_\theta
  = \mathcal{C}_{\mathrm{out}}
  \circ \Bigl(\prod_{i=1}^{L_h}\mathcal{C}_{h,i}\Bigr)
  \circ \Bigl(\prod_{i=1}^{L_d}\mathcal{C}_{d,i}\Bigr)
  \circ \mathcal{C}_{\mathrm{in}}.
\end{equation}
For $\vec x \in \mathbb{R}^{N_{\mathrm{in}}}$ the individual layers are
\begin{align}
    \mathcal{C}_{\mathrm{in}} 
    &:\mathbb{R}^{N_{\mathrm{in}}} \rightarrow \mathbb{R}^{N_{\mathrm{width}}}, 
    & \mathcal{C}_{\mathrm{in}}(\vec x) 
    &:= \sigma \left( \vec M_{\mathrm{in}, \theta} \vec x + \vec b_{\mathrm{in}, \theta} \right), 
    \\
    \mathcal{C}_{\mathrm{d, i}} 
    &:\mathbb{R}^{N_{\mathrm{width}}} \rightarrow \mathbb{R}^{N_{\mathrm{width}}}, 
    & \mathcal{C}_{\mathrm{d, i}}(\vec x) 
    &:=\sigma \bra{ \vec B_{p_\mathrm{drop}} \vec M_{\mathrm{d, i}, \theta} \vec x + \vec b_{\mathrm{d, i}, \theta} },
    \\
    \mathcal{C}_{\mathrm{h, i}} 
    &:\mathbb{R}^{N_{\mathrm{width}}} \rightarrow \mathbb{R}^{N_{\mathrm{width}}}, 
    & \mathcal{C}_{\mathrm{h, i}}(\vec x) 
    &:= \sigma \bra{\vec M_{\mathrm{h, i}, \theta} \vec x + \vec b_{\mathrm{h, i}, \theta}},
    \\
    \mathcal{C}_{\mathrm{out}} 
    &:\mathbb{R}^{N_{\mathrm{width}}} \rightarrow \mathbb{R}^{M_1\times M_2 \times \ldots}, 
    & \mathcal{C}_{\mathrm{out}}(\vec x) 
    &:= \vec M_{\mathrm{out}, \theta}\vec x + \vec b_{\mathrm{out}, \theta} 
\end{align}
with $\sigma(\cdot)$ the ReLU activation and hidden width
$N_{\mathrm{width}}$. During training, dropout is applied to reduce
overfitting, at test time it remains active to generate stochastic
ensembles for uncertainty quantification.

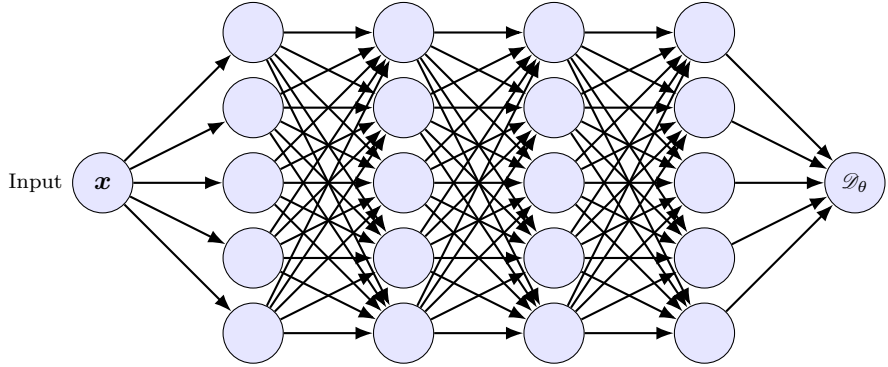
\begin{figure}[h!]
  \centering
  \input{nn.tex}
  \caption{Schematic of the surrogate network architecture in its 
  deterministic form. Input parameters $\vec x$ are passed through 
  stacked hidden layers with ReLU activation, producing a single 
  prediction $\mathscr{D}_\theta(\vec x)$. No dropout is applied at 
  test time, so repeated evaluations give identical outputs.}
  \label{fig:tikz_full_net}
\end{figure}

\begin{figure}[h!]
  \centering
  \input{nndp.tex}
  \caption{The same network evaluated with dropout active. At each 
  forward pass a Bernoulli mask randomly silences neurons, yielding a 
  stochastic prediction $\mathscr{D}^{(t)}_\theta(\vec x)$. Repeating 
  this process generates an ensemble from which predictive means and 
  variances are estimated, providing epistemic uncertainty.}
  \label{fig:tikz_drop_net}
\end{figure}
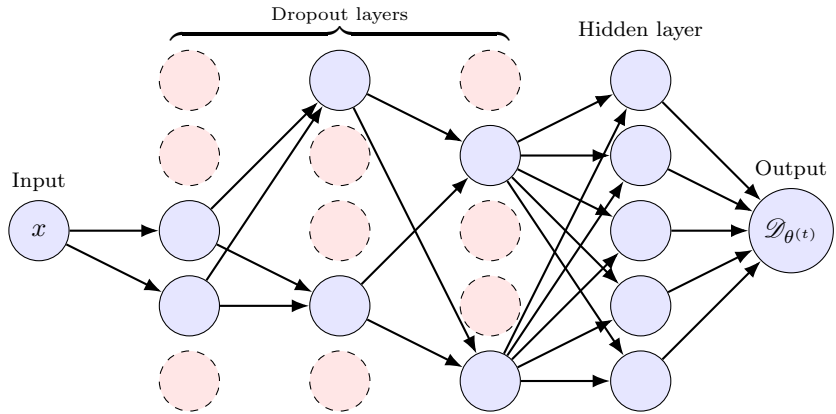

Network parameters are optimised by stochastic gradient descent to
minimise the quadratic loss
\begin{equation}
  \mathcal{L}(\theta)
  = \frac{1}{N}\sum_{i=1}^N
  \bigl\| \mathscr{D}_\theta(\vec x^{(i)}) - \vec d(\vec x^{(i)}) \bigr\|^2_{\ell^2}.
\end{equation}
Figures~\ref{fig:tikz_full_net} and \ref{fig:tikz_drop_net} illustrate
the architecture without and with dropout active. In practice, ReLU
activation provided the most stable training, smoother functions such
as softplus led to slower convergence.

\begin{algorithm}[H]
\caption{Training the surrogate network with dropout}
\label{alg:train_network}
\begin{algorithmic}[1]
\Require Training data $\mathrm{D}_{\mathrm{train}} = \{(\vec x^{(i)}, \vec d(\vec x^{(i)}))\}_{i=1}^N$, 
dropout probability $p_{\mathrm{drop}}$, learning rate $\eta$, 
number of iterations $N_{\mathrm{SGD}}$.
\State Initialise network parameters $\theta_0$
\For{$k = 0$ to $N_{\mathrm{SGD}}-1$}
  \State Sample a minibatch from $\mathrm{D}_{\mathrm{train}}$
  \State Apply dropout masks to form $\mathscr{D}_\theta^{(t)}$
  \State Update parameters $\theta_{k+1} \gets \theta_k - \eta \nabla_\theta \mathcal{L}(\theta_k)$
\EndFor
\end{algorithmic}
\end{algorithm}

\subsection{Monte Carlo dropout}

Dropout layers induce stochasticity in the network weights. At each
forward pass a Bernoulli mask is applied, yielding an effective
parameter vector $\theta^{(t)}$. Across passes $\{
\theta^{(t)}\}_{t=1}^T$ these parameters are independent draws from a
scaled Bernoulli distribution, which can be interpreted as approximate
sampling from the posterior $\mathbb{P}(\theta \,|\,
\mathrm{D}_{\mathrm{train}})$ \cite{gal2016dropout}. For a fixed input
$\vec x$, the predictive mean of the surrogate is then
\begin{equation}\label{eq:mc_mean}
\mathbb{E}\!\left[\mathscr{D}_{\theta}(\vec x)\right]
= \int_{\Theta} \mathscr{D}_{\theta}(\vec x)\,
   \mathbb{P}(\theta \,|\, \mathrm{D}_{\mathrm{train}})\,{\rm d}\theta
\;\approx\;
\frac{1}{T}\sum_{t=1}^T \mathscr{D}_{\theta^{(t)}}(\vec x),
\end{equation}
which is simply the ensemble average over $T$ stochastic forward
passes. The associated predictive variance is
\begin{equation}\label{eq:mc_var}
\begin{aligned}
\mathrm{Var}\!\left[\mathscr{D}_{\theta}(\vec x)\right]
&= \int_{\Theta}
  \bigl(\mathscr{D}_{\theta}(\vec x) -
         \mathbb{E}[\mathscr{D}_{\theta}(\vec x)]\bigr)^2\,
   \mathbb{P}(\theta \,|\, \mathrm{D}_{\mathrm{train}})\,{\rm d}\theta \\
&\approx \frac{1}{T-1}\sum_{t=1}^T
   \left(\mathscr{D}_{\theta^{(t)}}(\vec x) -
         \frac{1}{T}\sum_{t'=1}^T
         \mathscr{D}_{\theta^{(t')}}(\vec x)\right)^{\!2},
\end{aligned}
\end{equation}
where the square is understood elementwise across voxels. These
formulae provide practical estimators of the predictive mean and
pointwise variance, obtained by running the surrogate $T$ times with
dropout active. In this sense, Monte Carlo dropout mirrors the
structure of the original transport problem, just as MC dose
calculation estimates an expectation over random particle histories,
the surrogate estimates an expectation over random dropout masks.

\begin{algorithm}[H]
\caption{Training and uncertainty quantification with neural network surrogate}\label{alg:dropout_eval}
\begin{algorithmic}[1]
\Require Dropout probability $p_\mathrm{drop}$, number of MC passes $T$

\vspace{1mm}
\Statex \textit{// Offline training}
\State Define and train a neural network $\mathscr{D}_\theta$ as per algorithm \ref{alg:train_network}
\State Set $\theta \gets \theta_{\rm{best}}$ from algorithm \ref{alg:train_network}

\vspace{1mm}
\Statex \textit{// Online inference with uncertainty quantification}
\Function{PredictWithUncertainty}{$\vec x$}
    \State Set network to training mode to enable dropout
    \For{$t = 1$ to $T$}
        \State Sample $\vec x$ from distribution
        \State $\mathscr{D}_{\theta^{(t)}}(\vec x) \gets \mathscr{D}_\theta(\vec x)$ \Comment{Stochastic forward pass}
    \EndFor
    \State Compute predictive mean and variance using \eqref{eq:mc_mean}--\eqref{eq:mc_var}
    \State \Return $(\mathbb{E}[\mathscr{D}_{\theta}(\vec x)], \operatorname{Var}[\mathscr{D}_{\theta}(\vec x)])$
\EndFunction
\end{algorithmic}
\end{algorithm}

\subsection{Uncertainty sources and variance decomposition}

Predictions from the surrogate are random for two distinct reasons.
First, Monte Carlo dropout introduces stochasticity in the weights at
test time, yielding an epistemic (model) component. This term reflects
the fact that the network is trained on finite data with finite
capacity, it vanishes in the idealised limit of infinite data and
model size. Second, the physical inputs themselves are uncertain.
Material densities, geometrical parameters and in later examples beam
configurations are modelled as random variables. This variability
induces a parametric component, corresponding to the range of
clinically plausible scenarios.

Formally, let $\vec x \sim \pi$ denote the random input (domain and 
beam parameters) and let $\theta^{(t)}$ denote the random dropout mask 
applied at test time. For a fixed voxel or pixel index $j$, the 
surrogate prediction $\mathscr{D}_{\theta^{(t)}}(\vec x)_j$ is then a 
real-valued random variable on the product space of $(\vec x, 
\theta^{(t)})$. The law of total variance gives
\begin{equation}
\mathrm{Var}\!\left[\mathscr{D}_{\theta^{(t)}}(\vec x)_j\right]
=
\mathbb{E}_{\vec x\sim\pi}\!\left[
  \mathrm{Var}_{\theta^{(t)}}\!\left(
     \mathscr{D}_{\theta^{(t)}}(\vec x)_j \mid \vec x
  \right)\right]
+
\mathrm{Var}_{\vec x\sim\pi}\!\left(
  \mathbb{E}_{\theta^{(t)}}\!\left[
     \mathscr{D}_{\theta^{(t)}}(\vec x)_j \mid \vec x
  \right]\right).
\label{eq:ltv}
\end{equation}
The first term is the epistemic component, variance due to dropout at
fixed $\vec x$, averaged across possible inputs. The second term is
the parametric component, variance induced by sampling the input
parameters themselves. In later experiments we will estimate both
contributions numerically and report voxelwise maps as well as
aggregated summaries over regions of interest, providing a direct
comparison between model ignorance and input-driven variability.

\subsection{Finite-sample estimators}

In practice the expectations in \eqref{eq:mc_mean}–\eqref{eq:ltv} are
approximated by finite ensembles of stochastic forward passes and
finite collections of input samples. For a fixed input $\vec x_s$ and
$T$ dropout realisations $\{\theta^{(t)}\}_{t=1}^T$, the empirical mean
and variance at voxel $j$ are
\begin{equation}
\widehat{\mu}_j(\vec x_s)
= \frac{1}{T}\sum_{t=1}^{T} \mathscr{D}_{\theta^{(t)}}(\vec x_s)_j,
\qquad
\widehat{\sigma}^{2}_{\mathrm{drop},j}(\vec x_s)
= \frac{1}{T-1}\sum_{t=1}^{T}\bigl(\mathscr{D}_{\theta^{(t)}}(\vec x_s)_j-\widehat{\mu}_j(\vec x_s)\bigr)^2.
\label{eq:mu_sigma_drop}
\end{equation}

When inputs are also random, we draw $S$ independent samples
$\vec x^{(1)},\ldots,\vec x^{(S)}\sim\pi$ and average the estimators in
\eqref{eq:mu_sigma_drop} to obtain
\begin{equation}
\overline{\mu}_j
= \frac{1}{S}\sum_{s=1}^{S}\widehat{\mu}_j(\vec x^{(s)}),\qquad
\widehat{\mathrm{Var}}_{\mathrm{epi},j}
= \frac{1}{S}\sum_{s=1}^{S}\widehat{\sigma}^{2}_{\mathrm{drop},j}(\vec x^{(s)}),\qquad
\widehat{\mathrm{Var}}_{\mathrm{par},j}
= \frac{1}{S-1}\sum_{s=1}^{S}\bigl(\widehat{\mu}_j(\vec x^{(s)})-\overline{\mu}_j\bigr)^2.
\label{eq:var_epi_par}
\end{equation}
The total predictive variance estimator is then
\begin{equation}
\widehat{\mathrm{Var}}_{\mathrm{tot},j}
=\widehat{\mathrm{Var}}_{\mathrm{epi},j}+\widehat{\mathrm{Var}}_{\mathrm{par},j},
\label{eq:var_tot}
\end{equation}
providing a plug-in approximation of the decomposition in
\eqref{eq:ltv}.

For evaluation, we use two 
diagnostics. First, voxelwise maps
$j\mapsto\widehat{\mathrm{Var}}_{\mathrm{epi},j}$ and
$j\mapsto\widehat{\mathrm{Var}}_{\mathrm{par},j}$ show the spatial
structure of epistemic and parametric components, with scalar
summaries reported over clinically relevant regions or depth slabs.
Second, reliability curves compare nominal versus empirical coverage
levels (50–95\%) using either the dropout ensemble alone or the joint
ensemble over $(\vec x,\theta^{(t)})$, thereby quantifying calibration.
In all numerical experiments we
report these quantities for representative test instances and as
aggregated statistics across the test set.
\section{Numerical experiments}
\label{sec:numerics}

The purpose of this section is to validate the proposed surrogate
pipeline across settings of increasing complexity. Starting from
controlled one-dimensional tests with analytic benchmarks, we build up
to two- and three-dimensional phantoms generated by high-fidelity
Monte Carlo. The one-dimensional experiments provide a clean
environment to establish convergence, variance decomposition, and
calibration properties. The higher-dimensional phantoms then
demonstrate that the surrogate captures clinically relevant dose
features such as distal fall-off and heterogeneity effects, while
retaining tractable uncertainty quantification. Together these
experiments show both the mathematical soundness of the approach and
its potential value in medical physics applications where rapid,
uncertainty-aware dose evaluation is needed.

\subsection{Foundational 1D experiments}
We begin with one-dimensional analytic benchmarks that provide a
controlled setting for proof of concept. Here the surrogate $\vec x
\mapsto \mathscr{D}_{\theta}(\vec x)$ is trained to reproduce
depth–dose curves from simplified transport models \cite{ashby2025efficient}, where the input
vector $\vec x$ encodes material and beam parameters. These examples enable quantification of accuracy, decomposition of variance 
into epistemic and parametric components, and test empirical coverage of
dropout-based intervals against analytic ground truth. Establishing these properties in 1D provides a baseline
before extending to more realistic, higher-dimensional phantoms.

\subsection*{Example 1: 1-D analytic benchmark}

In this example, the input
vector comprises four parameters,
\begin{equation}
    \vec x = (\alpha, p, \rho, E_{\mathrm{peak}}),
\end{equation}
where $\alpha$ and $p$ are the Bragg-Kleeman parameters for the
medium, $\rho$ is the material density, and $E_{\mathrm{peak}}$ is the
peak energy at the inflow boundary. Perturbations in these inputs primarily shift the location of the Bragg peak. Direct averaging of depth–dose curves across samples consequently flattens the distal edge and obscures meaningful structure. To separate range from shape, we introduce two
surrogate tasks. A scalar range model
$\mathscr{R}_\theta:\mathbb{R}^4\to\mathbb{R}$ that predicts the
distal edge, and a shape model
$\mathscr{D}_\theta:\mathbb{R}^4\to\mathbb{R}^M$ that predicts the
curve on a uniform grid $\{z_j\}_{j=1}^M$ up to
$\mathscr{R}_\theta(\vec x)$.

The phantom consists of a homogeneous $20$cm water slab. The incident spectrum
at $z=0$ is Gaussian with mean $E_{\mathrm{peak}}$ and variance
$3.0$. Uncertainty is applied to the mean $E_{\mathrm{peak}}$, rather than to the distribution itself. Input uncertainties are modelled as
\begin{equation}\label{eq:distribution_of_x_values}
\begin{aligned}
    \alpha &\sim N(0.00246,0.000128), &
    p &\sim N(1.75,0.0102), \\
    \rho &\sim N(1.0,0.01), &
    E_{\mathrm{peak}} &\sim N(130.0,5.0).
\end{aligned}
\end{equation}
The distributions for $\alpha,p,\rho$ are informed by comparisons of
three Bragg-Kleeman parameterisations
\cite{pettersen2018accuracy,boon1998dosimetry,bortfeld1997analytical}.

We generated $N=1000$ phantoms and trained both the range and shape models
with identical hyperparameters: $L_h=3$ hidden layers, $L_d=3$ dropout
layers, hidden width $N_{\mathrm{width}}=512$, dropout probability
$p_{\mathrm{drop}}=0.05$, learning rate $\eta=10^{-3}$, and AdamW
optimisation for $3000$ epochs. The only difference is the output
dimension. During evaluation, we used $T=10^3$ dropout passes for the
shape model and $T=10^5$ for the range model.

\begin{figure}[h!]
\centering
\begin{subfigure}{0.49\textwidth}
    \includegraphics[width=\textwidth]{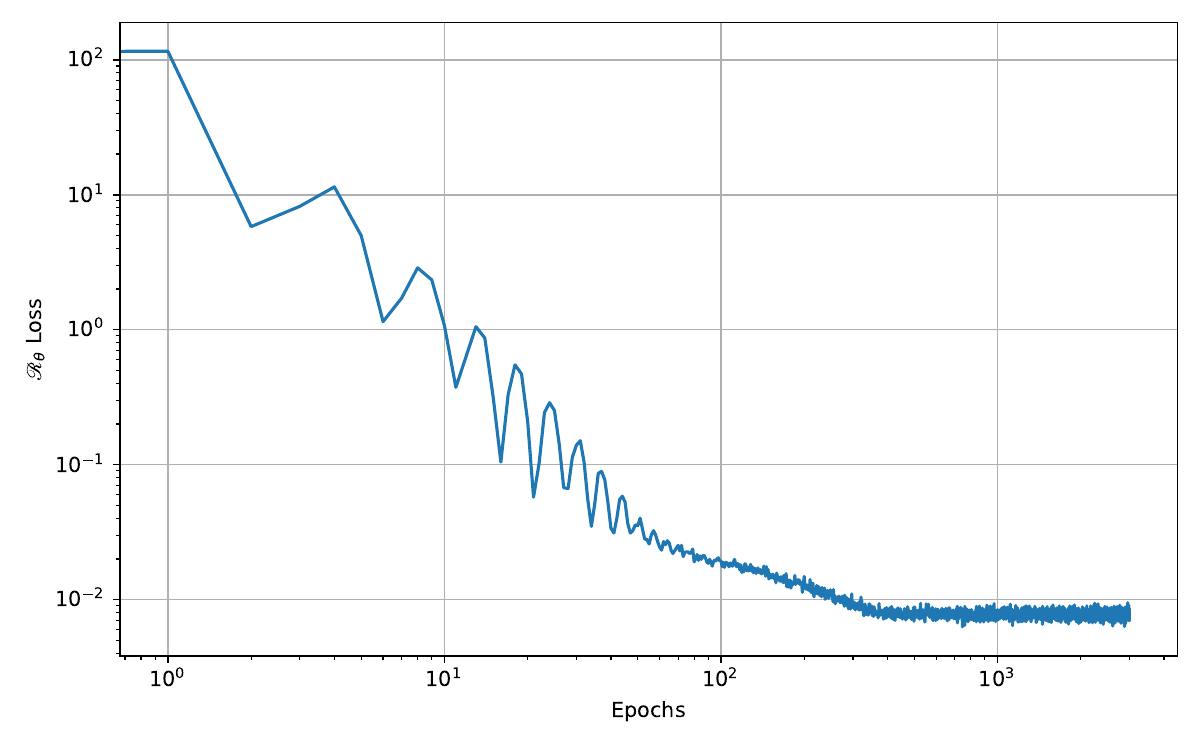}
    \caption{Training loss for the range model $\mathscr{R}_{\theta}$.}
    \label{fig:1D_range_loss_history}
\end{subfigure}
\hfill
\begin{subfigure}{0.49\textwidth}
    \includegraphics[width=\textwidth]{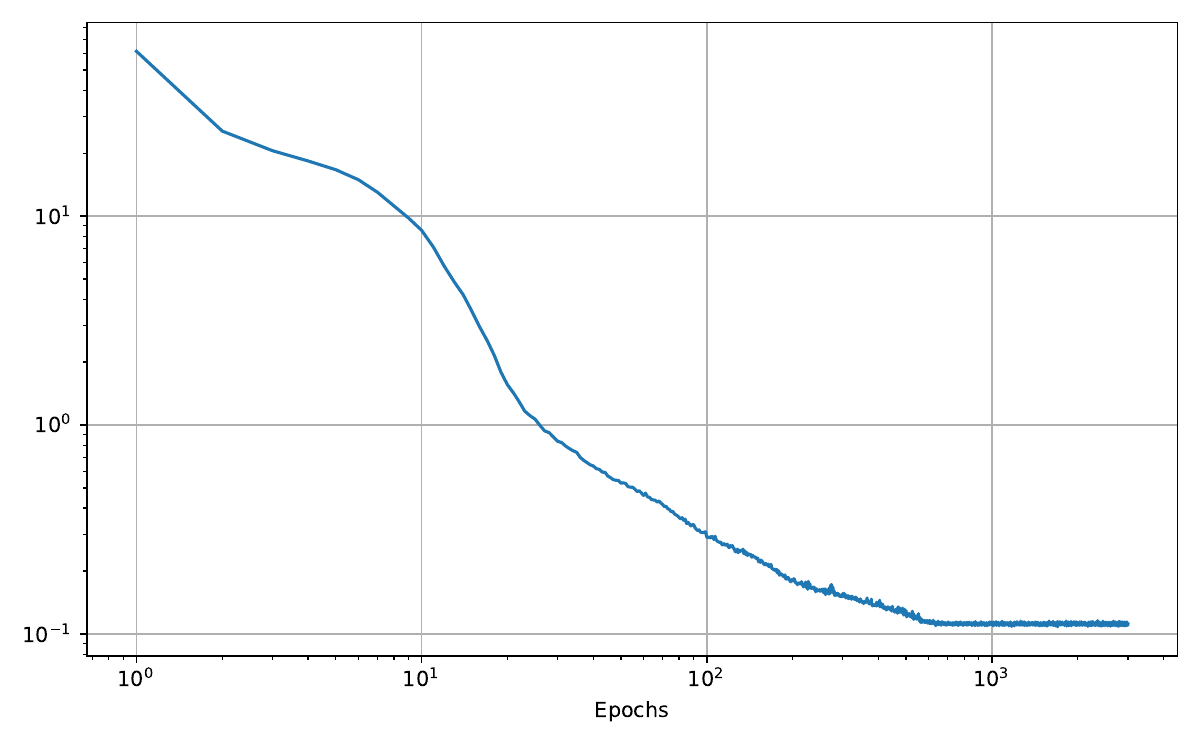}
    \caption{Training loss for the shape model $\mathscr{D}_{\theta}$.}
    \label{fig:1D_shape_loss_history}
\end{subfigure}
        
\caption{\textit{(Example 1)} Convergence of the surrogate models. The
  $\ell^2$ loss decreases steadily for both the range (left) and shape
  (right) networks, indicating stable training.}
\label{fig:Experiment_1_1D_loss_history}
\end{figure}

\begin{figure}[htbp]
\centering
\begin{subfigure}{0.49\textwidth}
    \includegraphics[width=\linewidth]{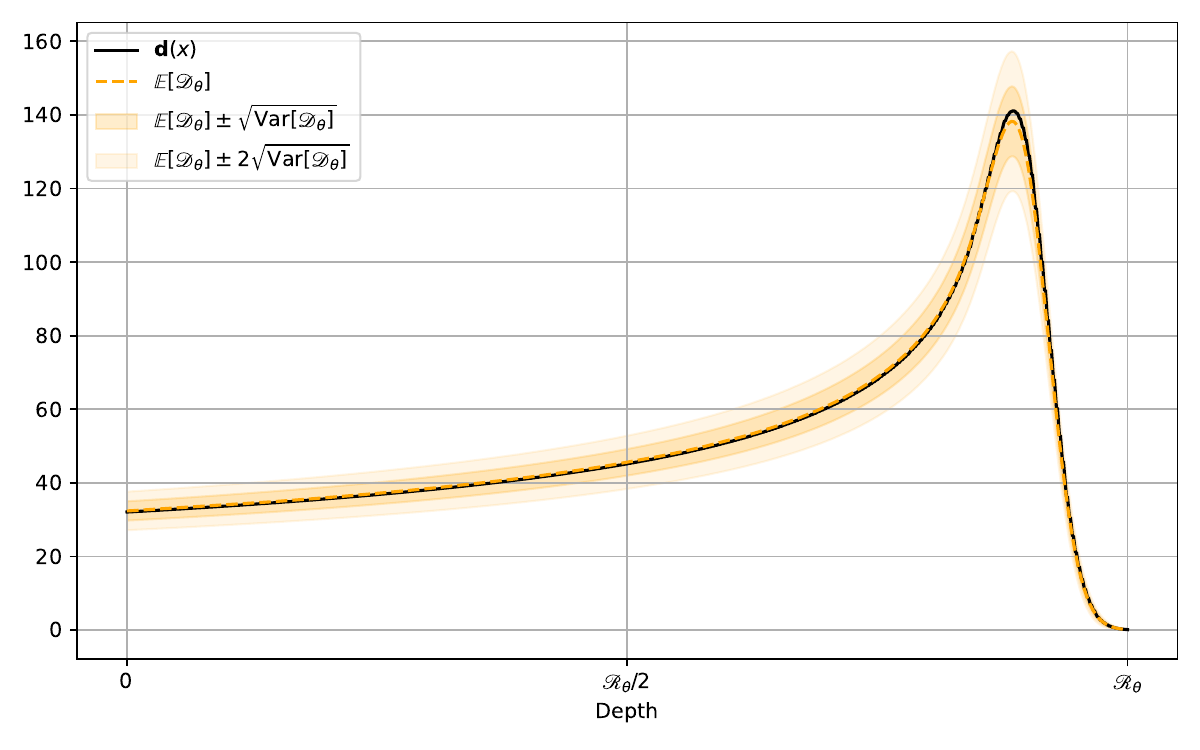}
    \caption{Predicted dose-depth curve from the
      shape model $\mathscr{D}_{\theta}$. The solid line is the
      ensemble mean, with shaded bands showing $\pm1$ and $\pm2$
      standard deviations. Variance localises around the Bragg peak
      and distal fall-off.}
    \label{fig:Experiment_1_shape_graph}
\end{subfigure}
\hfill
\begin{subfigure}{0.49\textwidth}
    \includegraphics[width=\linewidth]{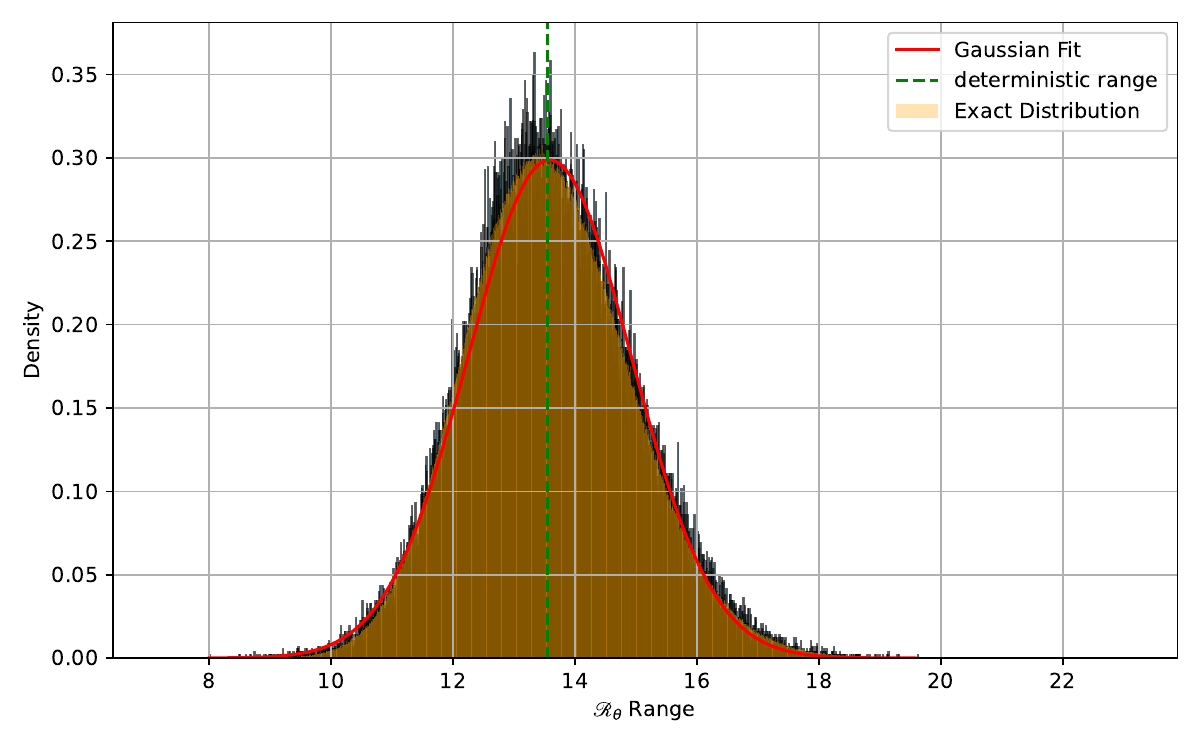}
    \caption{Distribution of predicted range
      values from the surrogate $\mathscr{R}_{\theta}$. The histogram
      is compared with the exact distribution, the deterministic range, and a Gaussian fit, showing close agreement.}
    \label{fig:Experiment_1_range_uncertainty_histogram}
\end{subfigure}   
\caption{\textit{(Example 1)} Output plots of the shape model (left) and range model (right).}
    \label{fig:Experiment_1_shape_graph_and_range_uncertainty_histogram}
\end{figure}

\begin{figure}[htbp]
\centering
\begin{subfigure}{0.48\textwidth}
    \includegraphics[width=\textwidth]{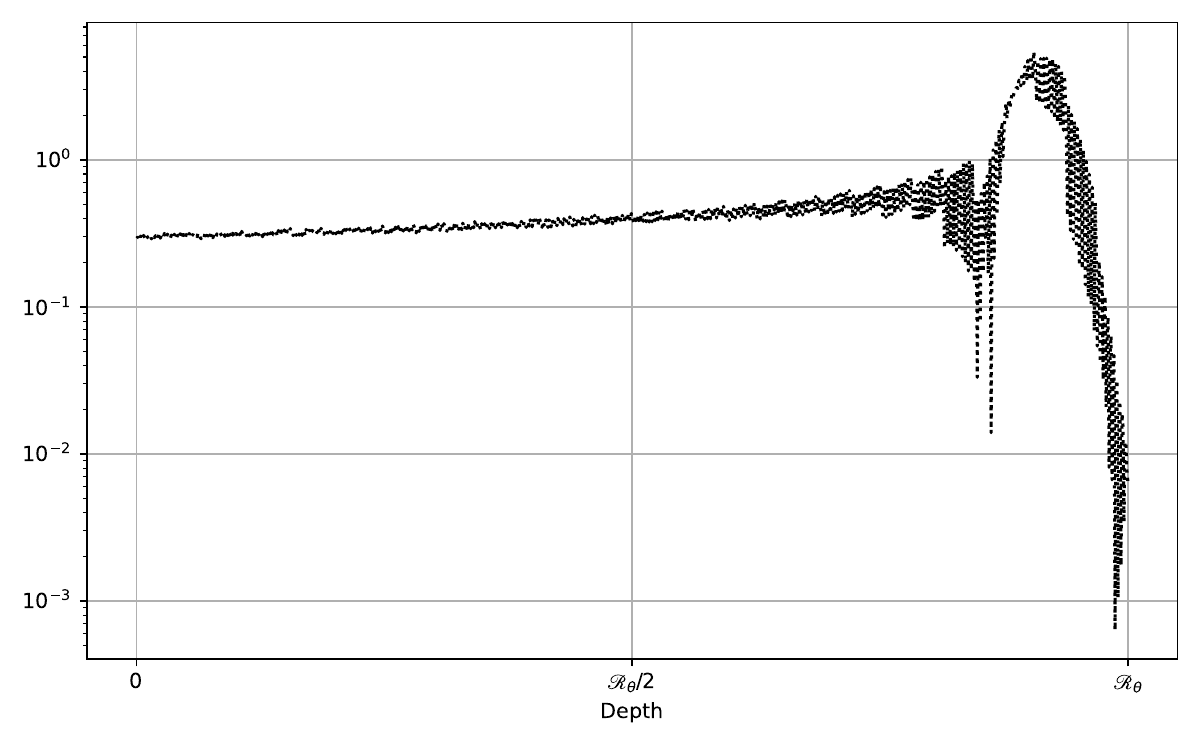}
    \caption{Absolute error $\abs{\vec d(\vec x) - \mathbb{E}[\mathscr{D}_{\theta}(\vec x)]}$.}
    \label{fig:Experiment_1_shape_absolute_pointwise_error}
\end{subfigure}
\hfill
\begin{subfigure}{0.48\textwidth}
    \includegraphics[width=\textwidth]{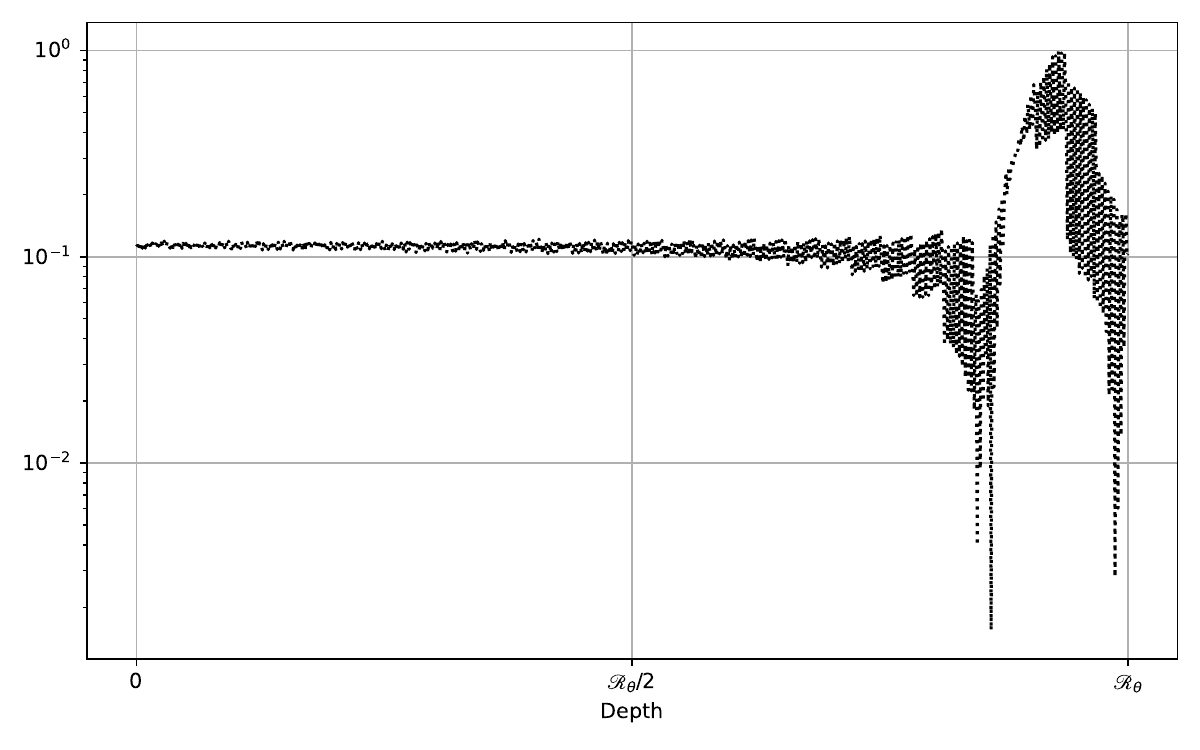}
    \caption{Normalised error $\abs{\vec d(\vec x) - \mathbb{E}[\mathscr{D}_{\theta}(\vec x)]}/\sqrt{\mathrm{Var}[\mathscr{D}_{\theta}(\vec x)]}$.}
    \label{fig:Experiment_1_shape_relative_pointwise_error}
\end{subfigure}
        
\caption{\textit{(Example 1)} Pointwise error of the shape model
  $\mathscr{D}_{\theta}$. Errors remain small across most depths, with the largest deviations occurring near the Bragg peak.}
\label{fig:Experiment_1_1D_pointwise_error}
\end{figure}

Figure~\ref{fig:Experiment_1_1D_loss_history} shows the loss history
for both surrogates, confirming convergence. In
Figure~\ref{fig:Experiment_1_shape_graph}, the shape model predictions
are plotted with $\pm1$ and $\pm2$ standard deviation bands. Variance
is small and tightly fitted along the proximal tail, increases around
the Bragg peak, and again narrows at the distal edge. The range model
distribution is summarised in
Figure~\ref{fig:Experiment_1_range_uncertainty_histogram}, where the
predicted distribution aligns well with the exact range and Gaussian
fit. Finally, Figure~\ref{fig:Experiment_1_1D_pointwise_error} shows
pointwise absolute and normalised errors between the surrogate and
exact data, demonstrating sub-percent agreement away from the distal
fall-off.

These results establish that the surrogate accurately reproduces the
analytic depth-dose model, while uncertainty localises in regions of
highest sensitivity such as the Bragg peak and distal edge.

\clearpage

\subsection*{Example 2: Convergence in training samples}

We next examine how the number of training samples $N$ affects
surrogate accuracy. Using the same range and shape models as in
Example~1, we compare two regimes: inputs drawn from the training
distribution \eqref{eq:distribution_of_x_values}, and a second in which the mean of the input distribution is shifted by two standard deviations.. 

\begin{figure}[h!]
\centering
\begin{subfigure}{0.49\textwidth}
    \includegraphics[width=\textwidth]{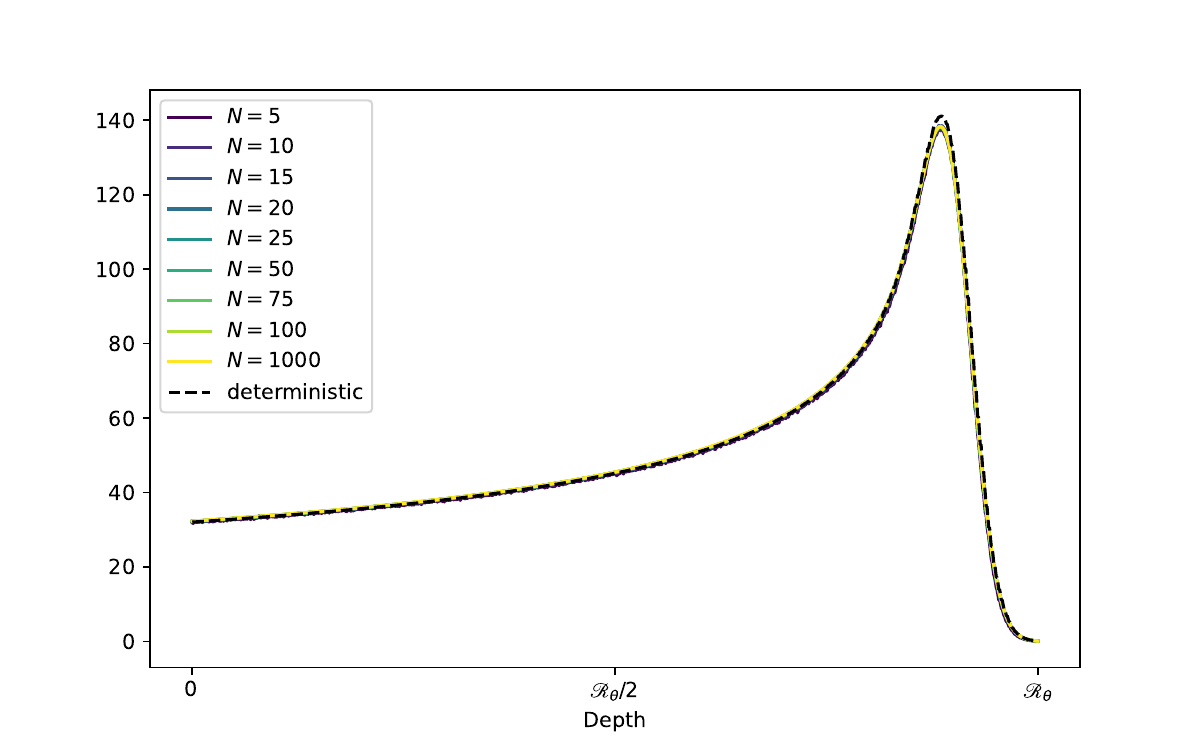}
    \caption{Expected dose shape, $\mathbb{E}[\mathscr{D}_{\theta}(\vec 0)]$.}
    \label{fig:Experiment_2_shape_Convergence_in_Training_set_size_mean}
\end{subfigure}
\hfill
\begin{subfigure}{0.49\textwidth}
    \includegraphics[width=\textwidth]{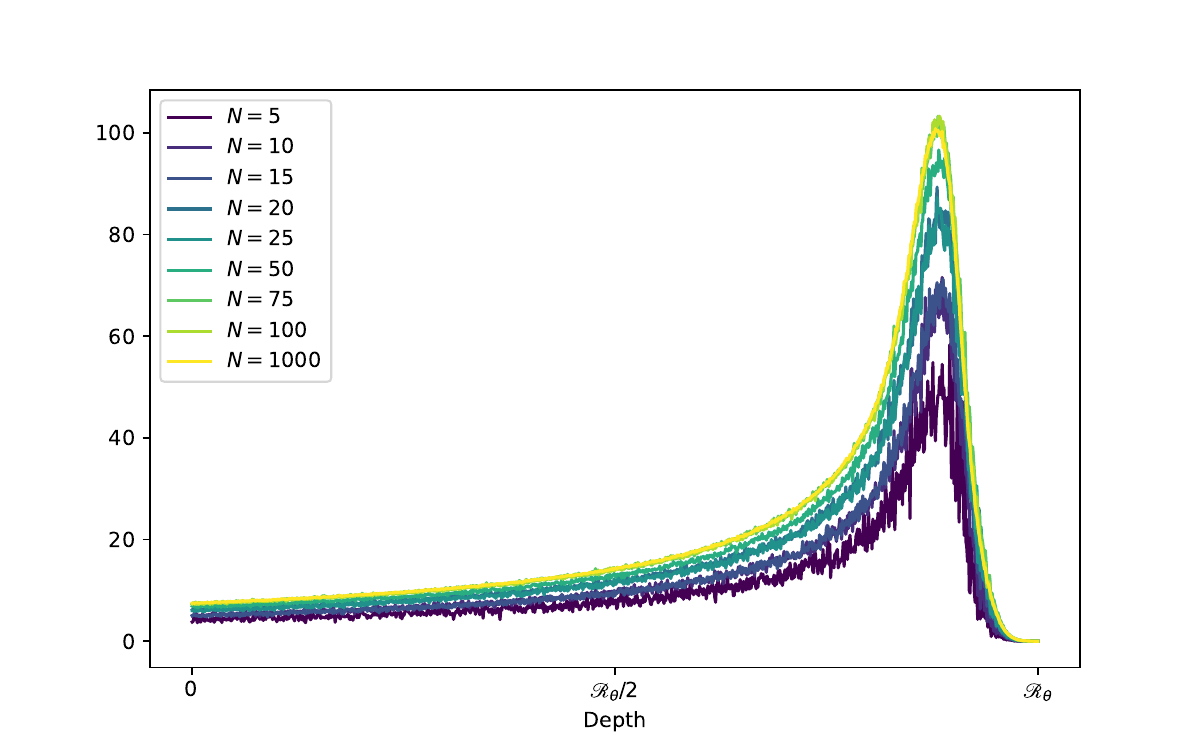}
    \caption{Dose variance, $\mathrm{Var}[\mathscr{D}_{\theta}(\vec 0)]$.}
    \label{fig:Experiment_2_shape_Convergence_in_Training_set_size_var}
\end{subfigure}
\hfill
\begin{subfigure}{0.49\textwidth}
    \includegraphics[width=\textwidth]{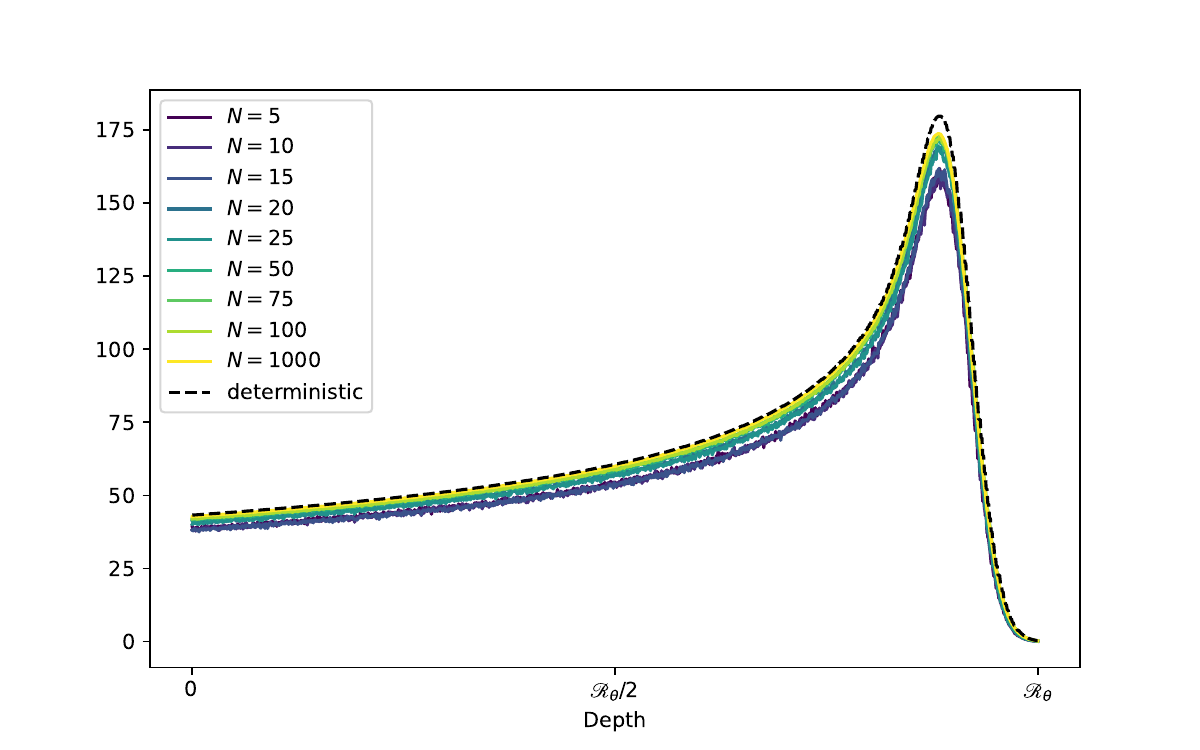}
    \caption{Expected dose shape, $\mathbb{E}[\mathscr{D}_{\theta}(-2\vec \sigma)]$.}
    \label{fig:Experiment_2_shape_Convergence_in_Training_set_size_mean_far}
\end{subfigure}
\hfill
\begin{subfigure}{0.49\textwidth}
    \includegraphics[width=\textwidth]{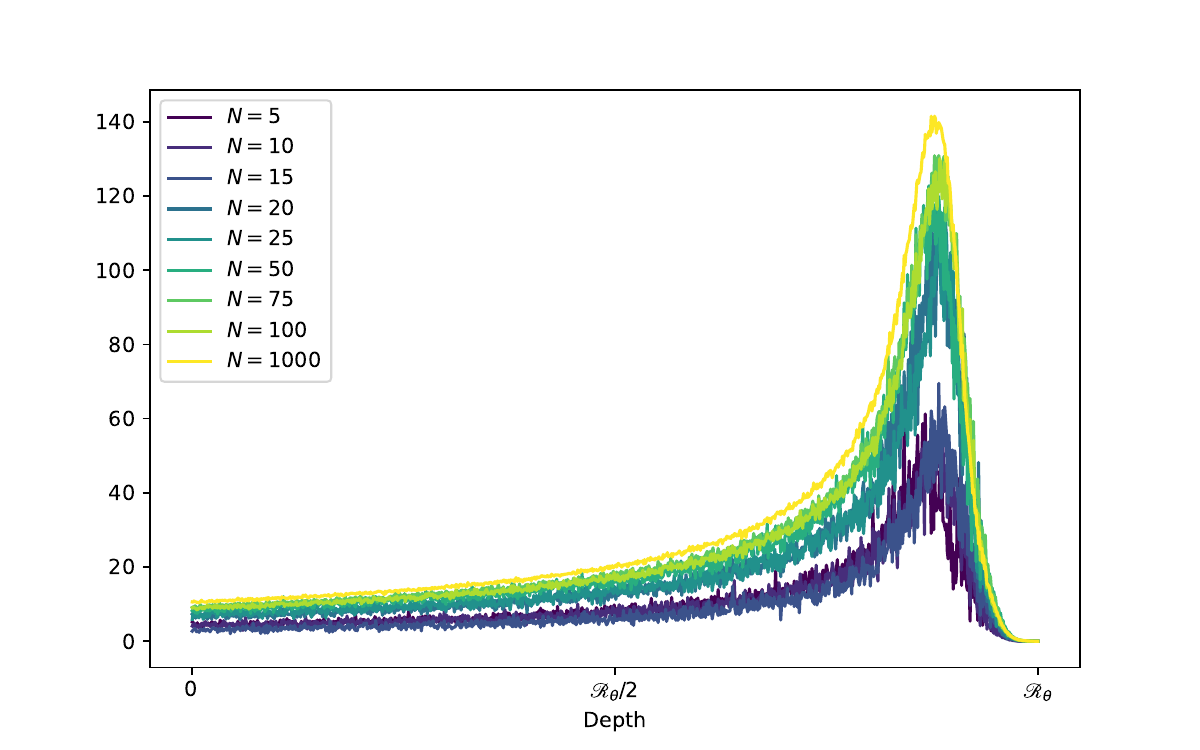}
    \caption{Dose variance, $\mathrm{Var}[\mathscr{D}_{\theta}(-2\vec \sigma)]$.}
    \label{fig:Experiment_2_shape_Convergence_in_Training_set_size_var_far}
\end{subfigure}
\caption{\textit{(Example 2)} Convergence of the shape model
$\mathscr{D}_\theta$ as the number of training samples $N$ increases.
Top: inputs drawn from the training distribution. Bottom: inputs with mean shifted by two standard deviations. Mean predictions are stable in-distribution, while out-of-distribution accuracy improves as $N$ increases.}
\label{fig:Experiment_2}
\end{figure}

For the shape model $\mathscr{D}_\theta$, Figure~\ref{fig:Experiment_2}
(top row) shows that within the training distribution the expected dose
(Figure~\ref{fig:Experiment_2_shape_Convergence_in_Training_set_size_mean})
and variance
(Figure~\ref{fig:Experiment_2_shape_Convergence_in_Training_set_size_var})
remain stable even for small values of $N$. By contrast, under the shifted
distribution the expected dose
(Figure~\ref{fig:Experiment_2_shape_Convergence_in_Training_set_size_mean_far})
and variance
(Figure~\ref{fig:Experiment_2_shape_Convergence_in_Training_set_size_var_far})
improve systematically with increasing $N$, reflecting the benefit of additional samples in covering previously unseen regions of parameter space.

\begin{figure}[h!]
\centering
\begin{subfigure}{0.49\textwidth}
    \includegraphics[width=\textwidth]{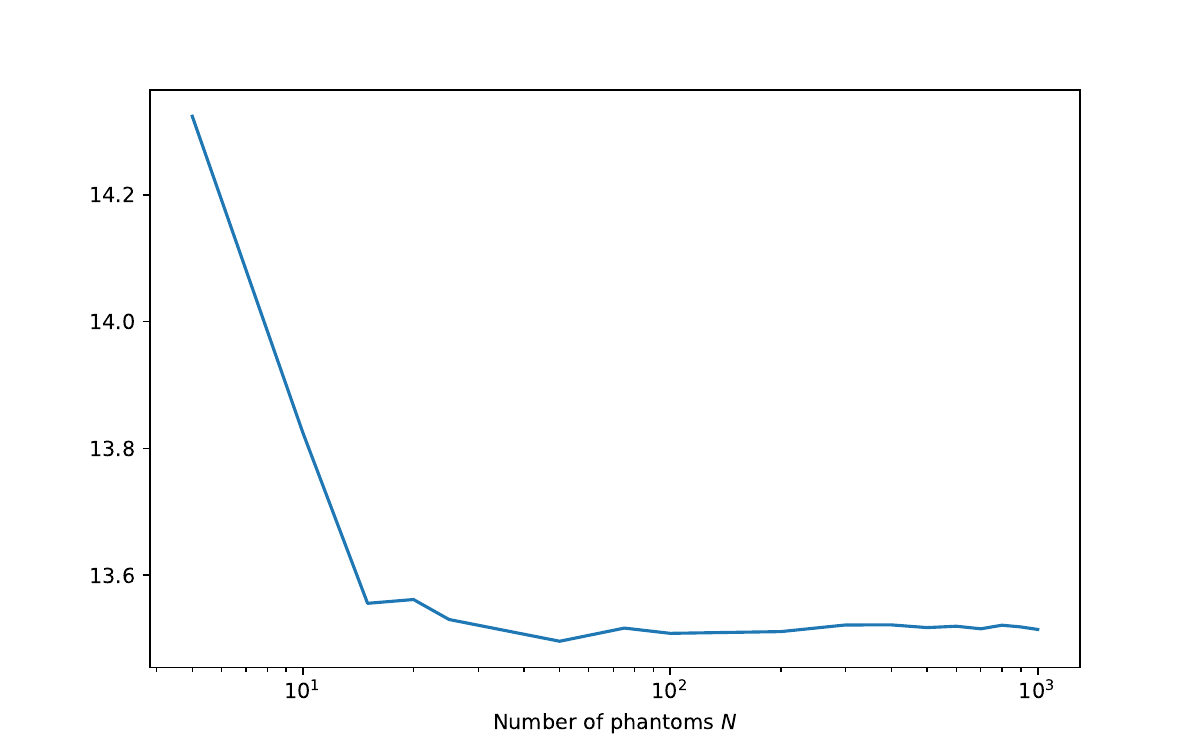}
    \caption{Expected range $\mathbb{E}[\mathscr{R}_{\theta}]$ against N.}
    \label{fig:Experiment_2_range_Convergence_in_Training_set_size_mean}
\end{subfigure}
\hfill
\begin{subfigure}{0.49\textwidth}
    \includegraphics[width=\textwidth]{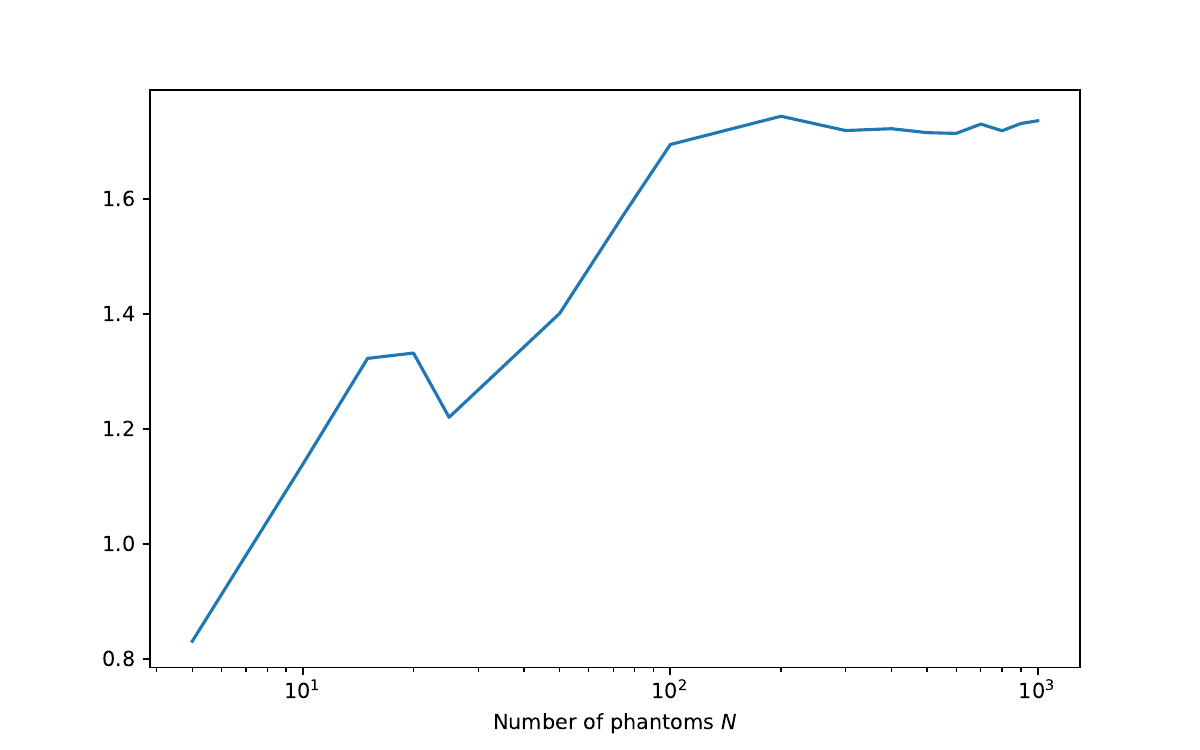}
    \caption{Range variance $\mathrm{Var}[\mathscr{R}_{\theta}]$ against N.}
    \label{fig:Experiment_2_range_Convergence_in_Training_set_size_var}
\end{subfigure}
\caption{\textit{(Example 2)} Convergence of the range model
$\mathscr{R}_\theta$. The expected range stabilises with
$N\approx 25$ samples, while the variance requires $N\approx 100$ to
converge.}
\label{fig:Experiment_2_range}
\end{figure}

For the range model $\mathscr{R}_\theta$, the same trend is observed in
Figure~\ref{fig:Experiment_2_range}. The expected range
(Figure~\ref{fig:Experiment_2_range_Convergence_in_Training_set_size_mean})
converges rapidly with as few as $N\approx 25$ samples, while the
variance
(Figure~\ref{fig:Experiment_2_range_Convergence_in_Training_set_size_var})
requires approximately $N\approx 100$ to stabilise.

\subsection*{Example 3: Convergence in MC-dropout passes}

Finally we assess how the number of Monte Carlo dropout passes $T$
affects stability of the estimators. For both the range and shape
models we compute predictive means and variances as $T$ increases.

\begin{figure}[h!]
\centering
\begin{subfigure}{0.49\textwidth}
    \includegraphics[width=\textwidth]{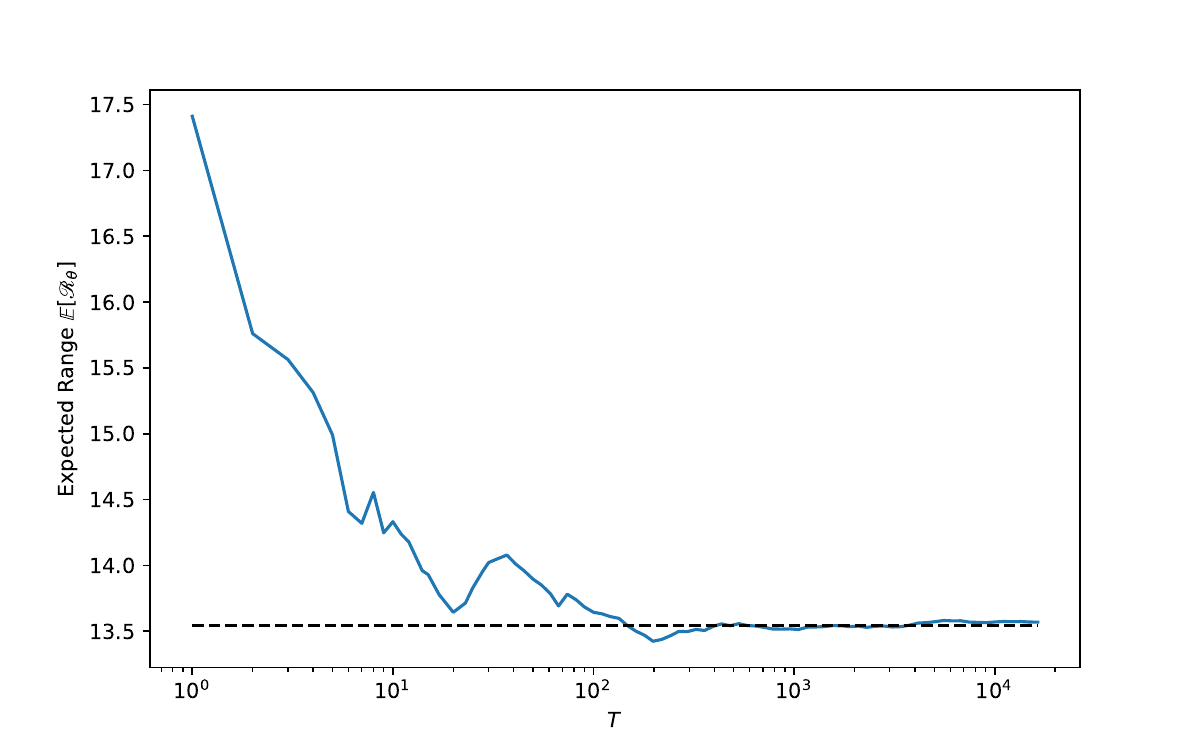}
    \caption{Expected range $\mathbb{E}[\mathscr{R}_{\theta}]$.}
    \label{fig:Experiment_3_range_Convergence_in_MC_dropout_passes_mean}
\end{subfigure}
\hfill
\begin{subfigure}{0.49\textwidth}
    \includegraphics[width=\textwidth]{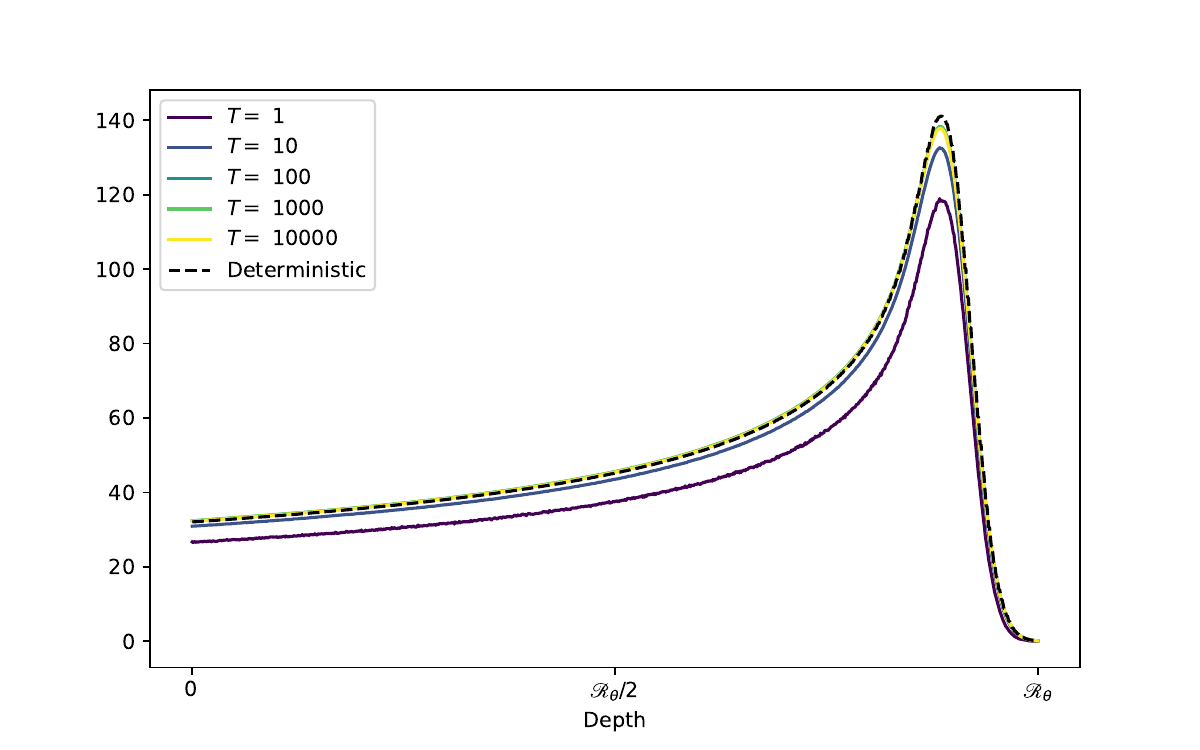}
    \caption{Expected shape $\mathbb{E}[\mathscr{D}_{\theta}]$.}
    \label{fig:Experiment_3_shape_Convergence_in_MC_dropout_passes_mean}
\end{subfigure}
\hfill
\begin{subfigure}{0.49\textwidth}
    \includegraphics[width=\textwidth]{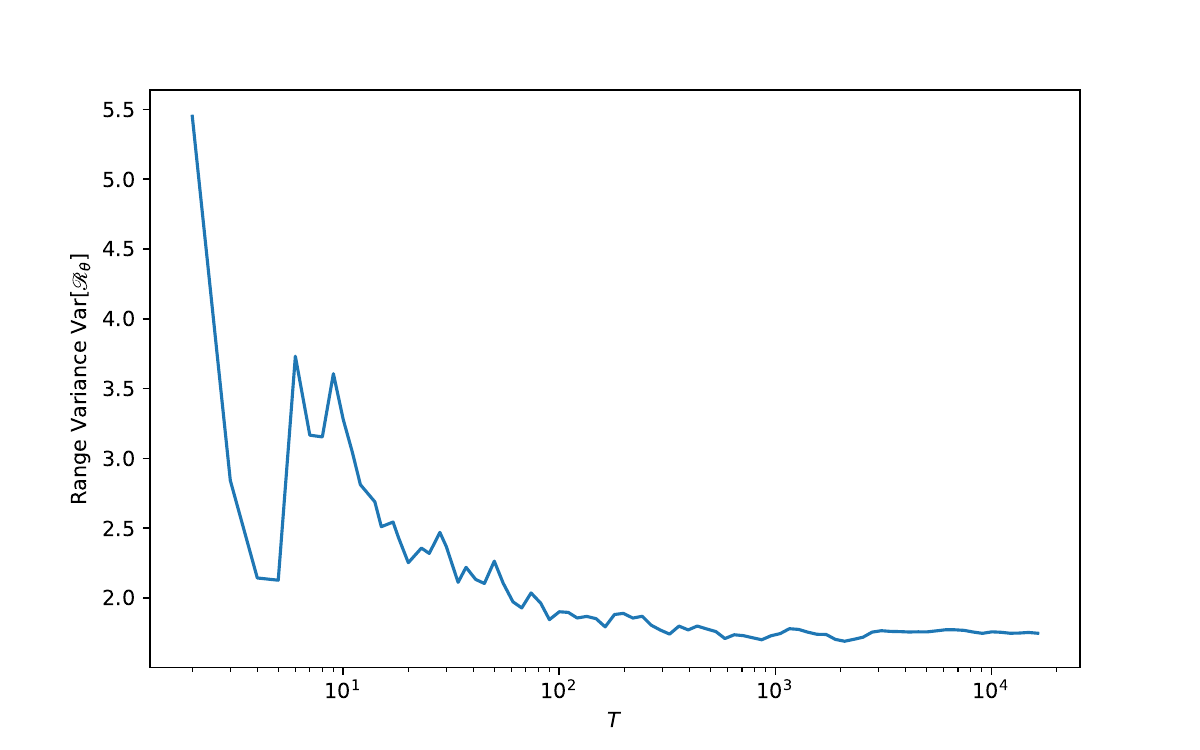}
    \caption{Variance of range $\mathrm{Var}[\mathscr{R}_{\theta}]$.}
    \label{fig:Experiment_3_range_Convergence_in_MC_dropout_passes_variance}
\end{subfigure}
\hfill
\begin{subfigure}{0.49\textwidth}
    \includegraphics[width=\textwidth]{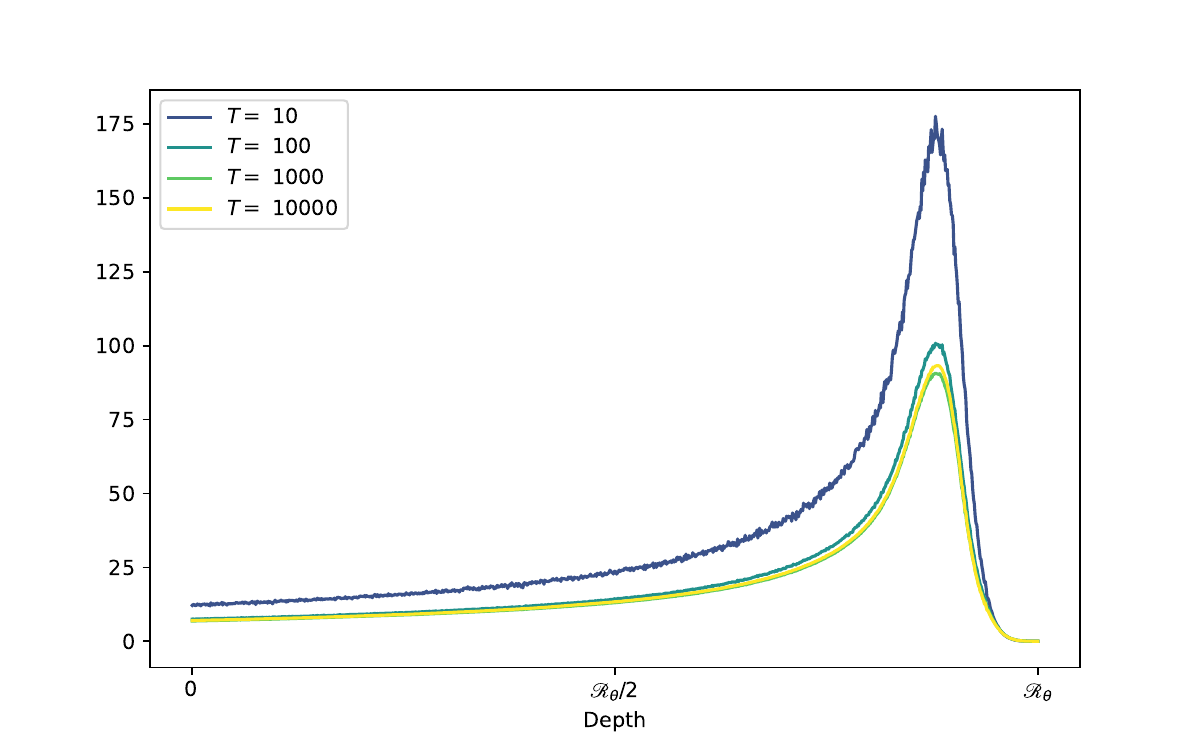}
    \caption{Variance of shape $\mathrm{Var}[\mathscr{D}_{\theta}]$.}
    \label{fig:Experiment_3_shape_Convergence_in_MC_dropout_passes_variance}
\end{subfigure}
\caption{\textit{(Example 3)} Convergence of mean and variance estimates
with the number of Monte Carlo dropout passes $T$. Mean predictions
stabilise quickly, while variance estimates decrease more gradually as
sampling noise is averaged out.}
\label{fig:Experiment_3}
\end{figure}

In Figure~\ref{fig:Experiment_3}, the expected range
(Figure~\ref{fig:Experiment_3_range_Convergence_in_MC_dropout_passes_mean})
and expected dose shape
(Figure~\ref{fig:Experiment_3_shape_Convergence_in_MC_dropout_passes_mean})
converge rapidly, indicating that relatively few dropout passes are
needed for stable mean predictions. In contrast, the variance
estimators
(Figures~\ref{fig:Experiment_3_range_Convergence_in_MC_dropout_passes_variance}–\ref{fig:Experiment_3_shape_Convergence_in_MC_dropout_passes_variance})
exhibit a gradual downward trend, reflecting reduced sampling error as
$T$ increases. This behaviour is expected, the empirical variance
converges more slowly than the empirical mean, and additional dropout
passes mainly reduce noise in the uncertainty estimates.

\clearpage

\subsection*{Example 4: Effects of dropout on epistemic uncertainty}

\begin{figure}[h!]
    \centering
    \includegraphics[width=0.75\linewidth]{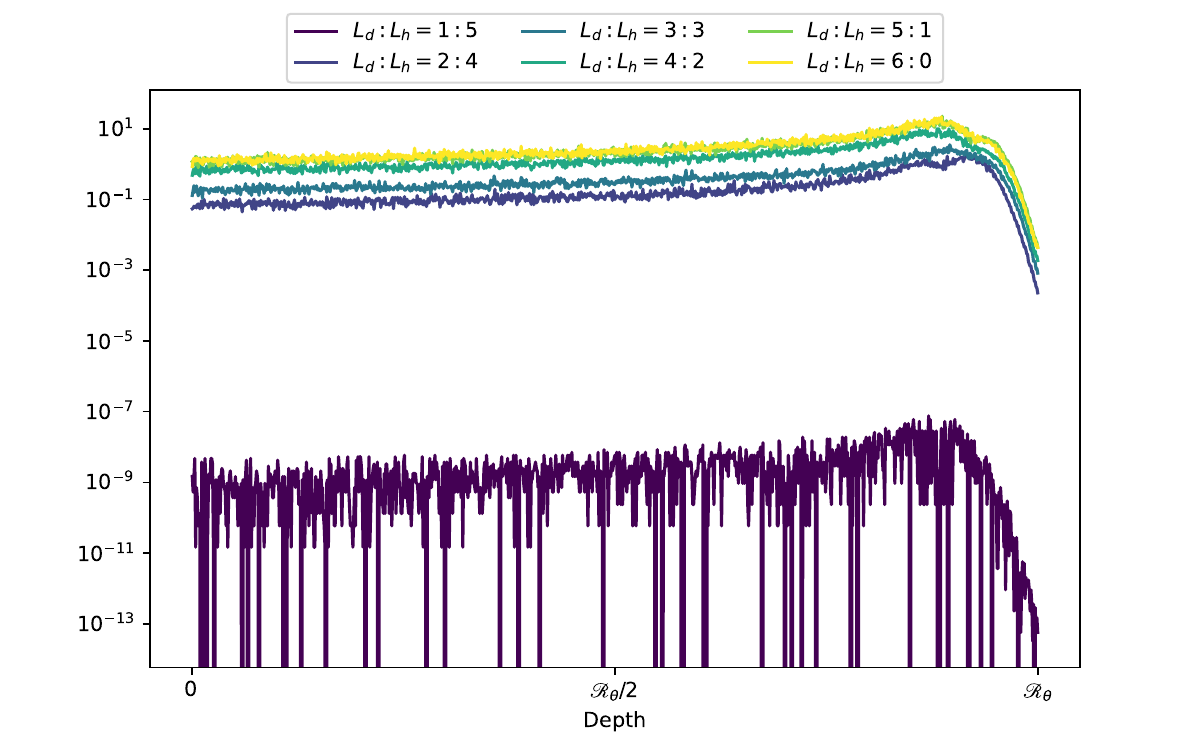}
    \caption{\textit{(Example 4)} Pointwise shape epistemic variance $\mathrm{Var}_{\mathrm{epi}}[\mathscr{D}_{\theta}(\vec x)]$ versus the ratio of dropout to hidden layers. More
    dropout layers systematically increase epistemic uncertainty.}
    \label{fig:Experiment_4_dropout_layers}
\end{figure}

We now examine how dropout design choices affect epistemic
uncertainty. Using the one-dimensional analytic model, we vary the
ratio of dropout layers to hidden layers $L_d : L_h$ while fixing the
dropout probability at $p_{\mathrm{drop}} = 0.05$. During evaluation we
set $\vec x$ equal to the mean values in
\eqref{eq:distribution_of_x_values}, rather than sampling from the
distribution, to isolate epistemic effects. As shown in
Figure~\ref{fig:Experiment_4_dropout_layers}, the pointwise epistemic
variance increases uniformly as dropout layers dominate.

\begin{figure}[h!]
\centering
\begin{subfigure}{0.40\textwidth}
    \includegraphics[width=\textwidth]{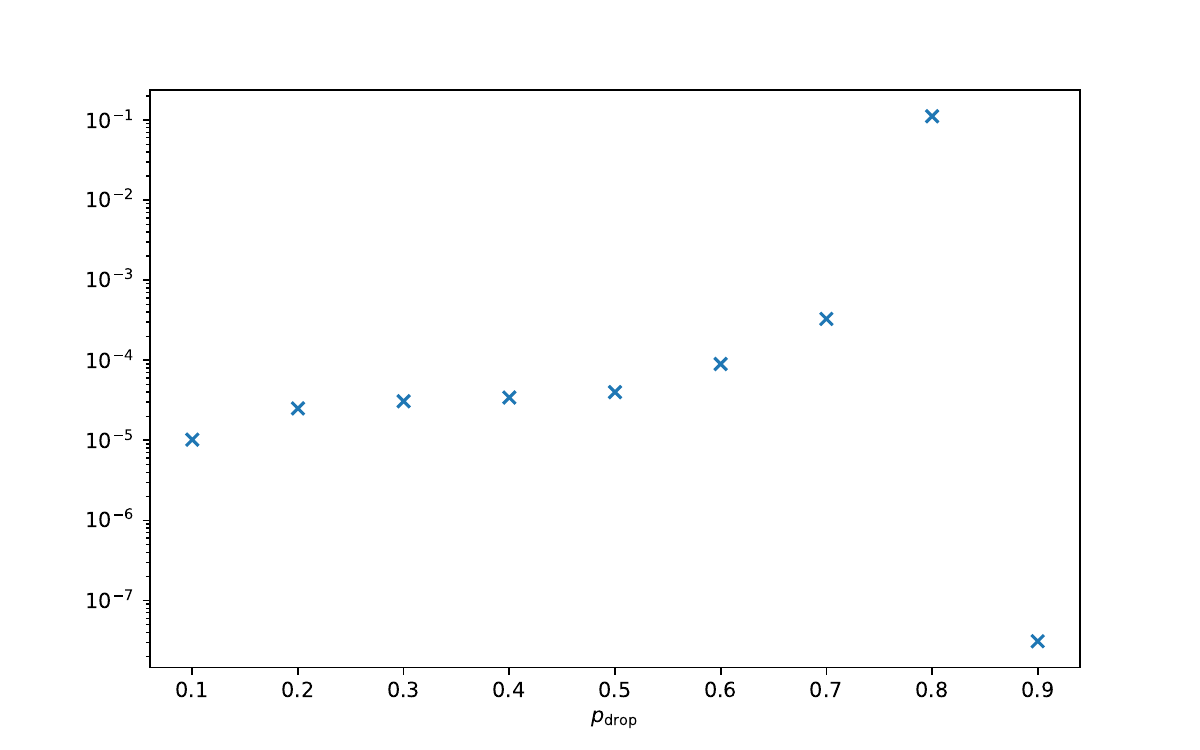}
    \caption{Range model: variance versus $p_{\mathrm{drop}}$ for $L_h,L_d=3$.}
    \label{fig:Experiment_4_dropout_range}
\end{subfigure}
\begin{subfigure}{0.40\textwidth}
    \includegraphics[width=\textwidth]{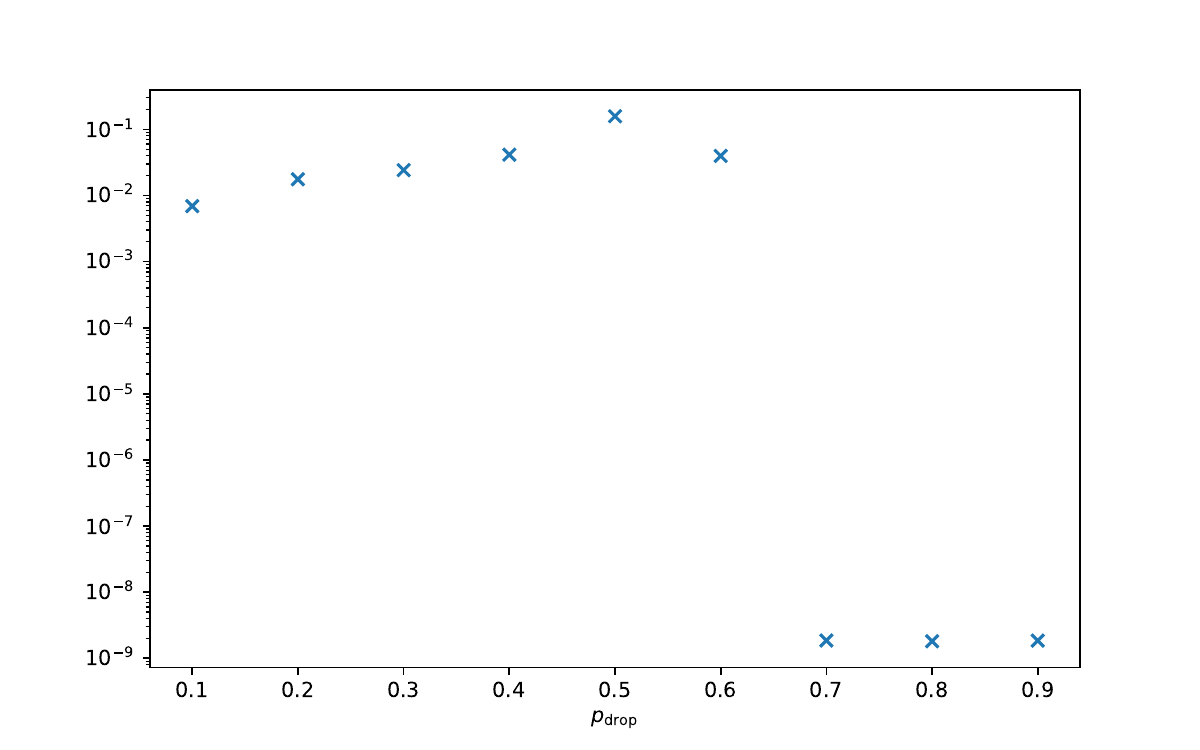}
    \caption{Shape model: variance versus $p_{\mathrm{drop}}$ for $L_h,L_d=0,6$.}
    \label{fig:Experiment_4_dropout_shape}
\end{subfigure}
\caption{\textit{(Example 4)} Effect of dropout probability on epistemic
variance. The range model shows a mild upward trend with instability
for large $p_{\mathrm{drop}}$, while the shape model remains flat until
a sharp increase near $p_{\mathrm{drop}}\approx 0.67$.}
\label{fig:Experiment_4_dropout_probability}
\end{figure}

In a second study, we fix the number of layer types and vary the
dropout probability $p_{\mathrm{drop}}$. For the range model
(Figure~\ref{fig:Experiment_4_dropout_range}), the average epistemic
variance increases slightly with $p_{\mathrm{drop}}$, with unstable
behaviour for very large values. For the shape model
(Figure~\ref{fig:Experiment_4_dropout_shape}), the variance is
remarkably consistent up to $p_{\mathrm{drop}}\approx 0.67$, beyond
which a sharp increase occurs, likely due to under-training when most
units are dropped. Note that for the shape model we set $L_d=6$ and
$L_h=0$; in early simulations with balanced ratios the variance was
robust to $p_{\mathrm{drop}}$.

\clearpage

\subsection{Higher-dimensional phantom studies}

Having established baseline behaviour in one-dimensional analytic
benchmarks, we now consider higher-dimensional experiments where the
surrogate is trained directly on Monte Carlo simulations of voxelised
phantoms. These cases move beyond simplified curves to data that more
closely resemble clinical dose distributions, with geometric
heterogeneity and beam-parameter variability.

The two-dimensional bone–water phantom provides a controlled setting
to probe uncertainty around material interfaces and distal fall-off,
while the three-dimensional water phantom demonstrates scalability to
volumetric outputs and realistic beam perturbations. In both settings
we decompose epistemic and parametric variance, examine calibration,
and assess behaviour under distribution shift. These experiments
illustrate the surrogate's performance under clinically motivated
conditions and its potential as a fast, uncertainty-aware alternative
to direct Monte Carlo evaluation.

\subsection*{Example 5: Two-dimensional bone–water phantom}

We now train the surrogate on Monte Carlo simulations of a
two-dimensional bone–water phantom, obtained by integrating
three-dimensional dose along the $z$-axis. This setting introduces
geometric heterogeneity while remaining computationally tractable.

The phantom is a cube 
\[
(-7.5,7.5)\times(-5,5)\times(-5,5)~\mathrm{cm}^3
\]
partitioned into a central bone slab surrounded by water. The bone
region is perturbed according to
\begin{equation}\label{eq:example_domain_perturb}
(-2.5+x_1-x_2,\,2.5+x_1+x_2)\times(-5,5)\times(-5,5)~\mathrm{cm}^3,
\qquad x_1,x_2\sim\mathrm{N}(0,0.1),
\end{equation}
where $x_1$ controls the position and $x_2$ the thickness. Dose is
simulated with TOPAS/Geant4 using $2.5\times10^5$ particle histories
per phantom and $N=50$ independent phantoms. The beam is a narrow
pencil-like Gaussian with spatial, angular and energy spreads of
$10^{-11}$\,cm, $10^{-10}$\,rad, and 1\,MeV respectively. The resulting
dose is integrated along $z$, shifted by $10^{-10}$ to avoid zero
values, and transformed with $\log_{10}$ to stabilise training. Each
sample is therefore a log-dose matrix $\vec d\in[-10,\infty)^{M_1\times M_2}$ with
resolution $M_1=1500$, $M_2=200$.

\begin{figure}[h!]
    \centering
    \includegraphics[width=0.75\linewidth]{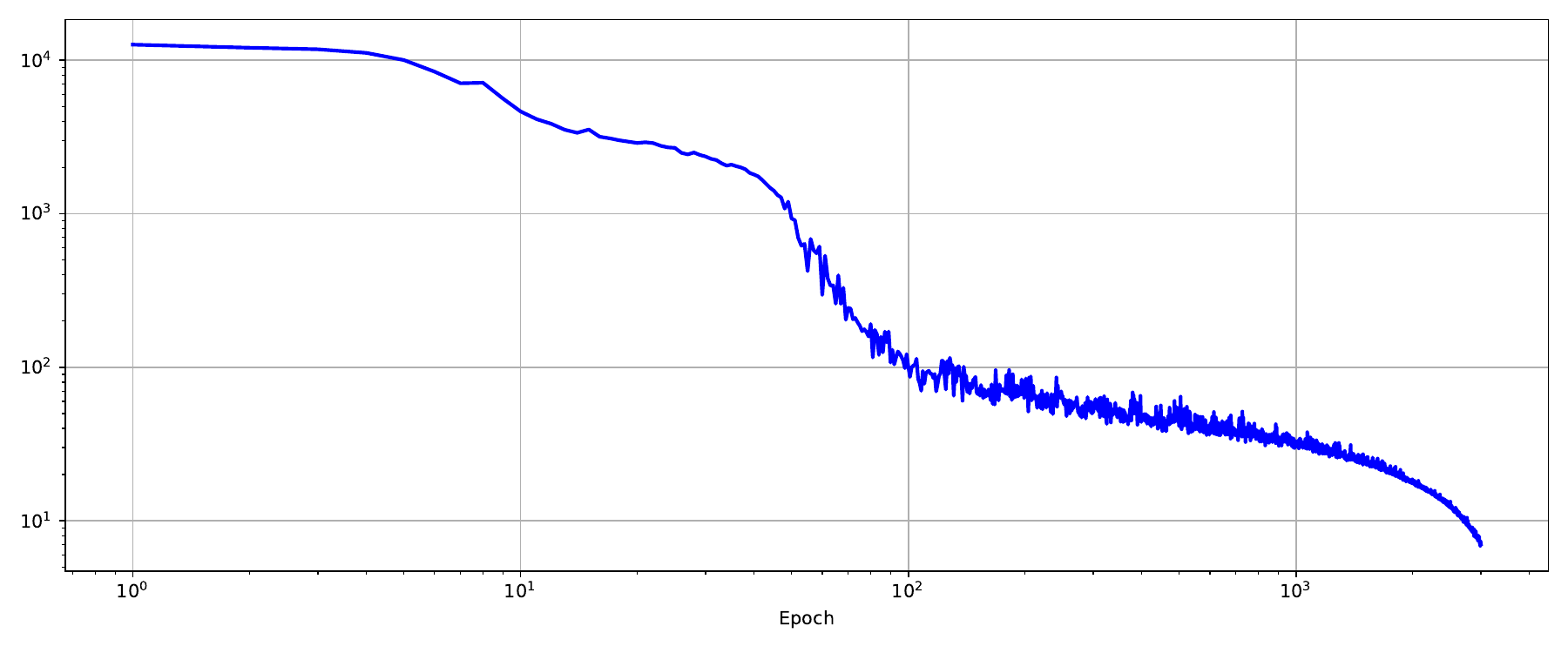}
    \caption{\textit{(Example 5)} Training history of the 2D surrogate.
    The $\ell^2$ loss between surrogate predictions $\mathscr{D}_\theta$
    and reference log-dose $\vec d$ decreases steadily over epochs.}
    \label{fig:loss}
\end{figure}

We train a network with $L_h=3$ hidden layers and $L_d=3$ dropout
layers, width $N_{\mathrm{width}}=512$, dropout probability
$p_{\mathrm{drop}}=0.05$, learning rate $\eta=10^{-3}$, and AdamW
optimisation. The network outputs $\log_{10}$ dose predictions with a
minimum of $-10$. Figure~\ref{fig:loss} shows the $\ell^2$ loss
history, confirming stable convergence.

\begin{figure}[h!]
    \centering
    \includegraphics[width=0.75\linewidth]{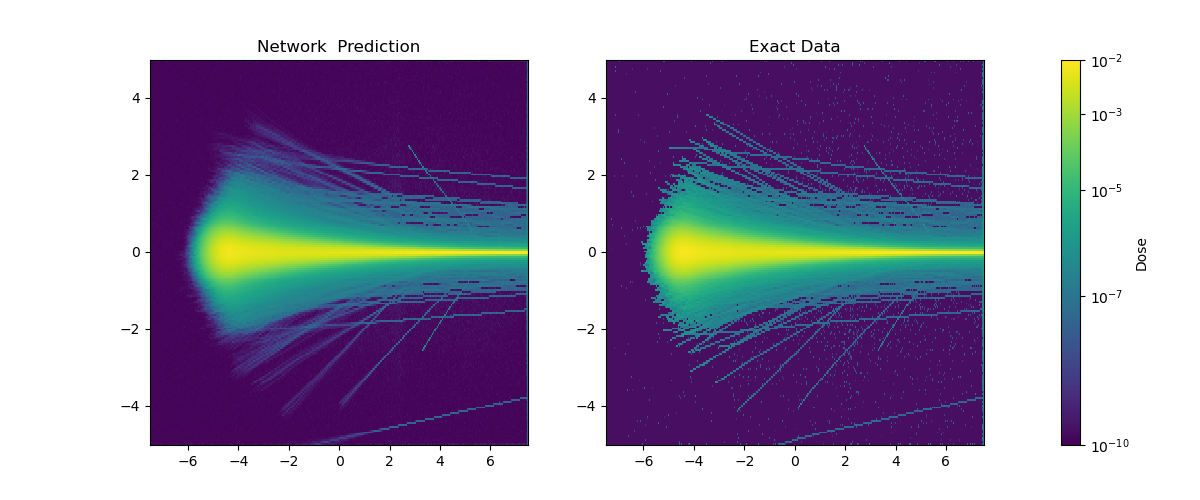}
    \caption{\textit{(Example 5)} Expected log-dose from the surrogate
    $\mathbb{E}[\mathscr{D}_\theta]$ (left) compared to Monte Carlo
    $\vec d$ (right) for $\vec x=(0,0)$. The surrogate reproduces the
    distal fall-off and lateral spread.}
    \label{fig:2D_plots1}
\end{figure}

\begin{figure}[h!]
\centering
\begin{subfigure}{0.40\textwidth}
    \includegraphics[width=\textwidth]{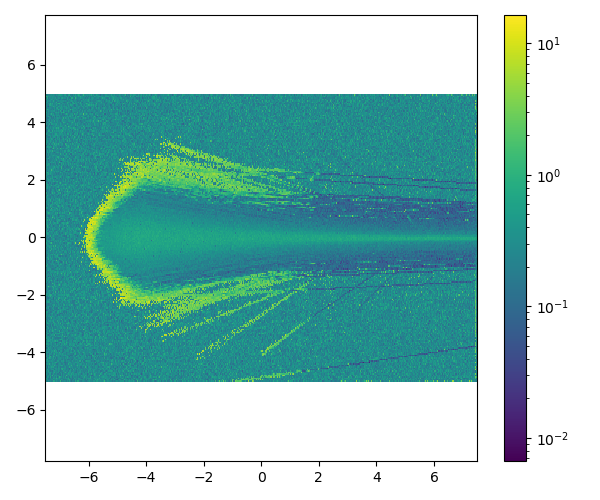}
    \caption{$\mathrm{Var}_{\mathrm{para}}[\mathscr{D}_{\theta}(\vec x)]$ .}
    \label{fig:2D_plots2_aleatoric}
\end{subfigure}
\begin{subfigure}{0.40\textwidth}
    \includegraphics[width=\textwidth]{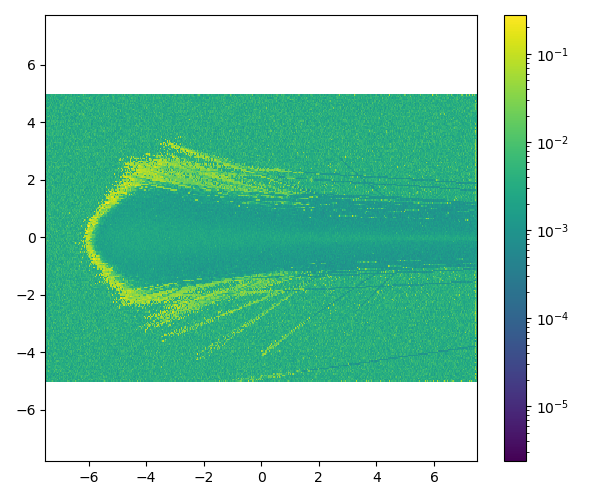}
    \caption{ $\mathrm{Var}_{\mathrm{epi}}[\mathscr{D}_{\theta}(\vec x)]$.}
    \label{fig:2D_plots2_epistemic}
\end{subfigure}
\caption{\textit{(Example 5)} Variance decomposition for $\vec x=(0,0)$, with parametric (left) and epistemic (right).
Uncertainty peaks along the distal edge and at the perturbed
bone–water boundary. Parametric variance dominates, while epistemic
variance remains localised.}
\label{fig:2D_2D_Input_standard_deviation}
\end{figure}

\begin{figure}[h!]
    \centering
    \includegraphics[width=0.75\linewidth]{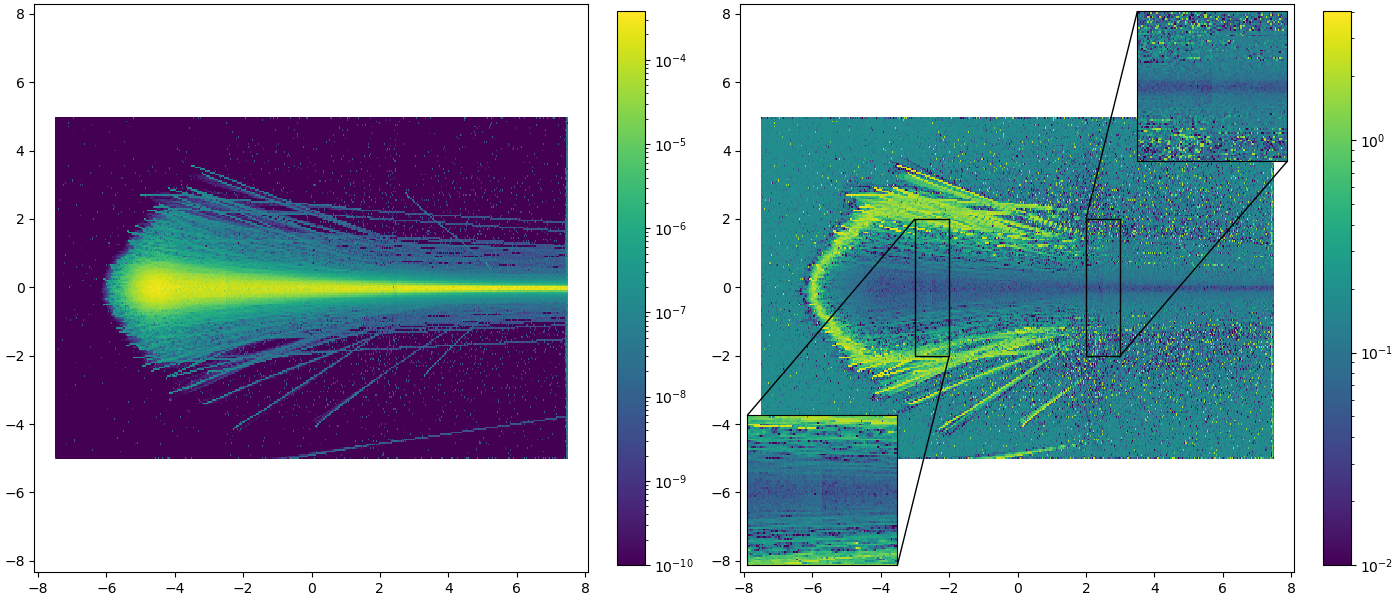}
    \caption{\textit{(Example 5)} Error maps for $\vec x=(0,0)$. Left: $|10^{\vec d(\vec x)} - 10^{\mathscr{D}(\vec x)}|$
    absolute error between surrogate and dose. Right: $|\vec d(\vec x) - \mathscr{D}(\vec x)|$ logarithmic error
    highlighting discontinuities at the bone–water boundary. Errors
    concentrate in regions of high uncertainty.}
    \label{fig:2D_plots3}
\end{figure}

For a representative test input $\vec x=(0,0)$, the surrogate mean
prediction agrees closely with Monte Carlo
(Figure~\ref{fig:2D_plots1}), capturing beam spread and magnitude.
Uncertainty maps (Figure~\ref{fig:2D_2D_Input_standard_deviation})
reveal that variance concentrates along the distal edge and the
bone–water interface at $x\approx 2.5$. Decomposition shows that
parametric variance dominates, consistent with geometry perturbations
being the primary source of variability. Error maps
(Figure~\ref{fig:2D_plots3}) confirm that most discrepancies occur near
high-uncertainty regions, especially around material boundaries.

\clearpage

\subsection*{Example 6: Two-dimensional phantom with domain and beam uncertainty}

We now extend the previous 2D bone–water phantom by incorporating
uncertainty in both the domain and the incident beam. The input vector
is four-dimensional, $\vec x=(x_1,x_2,x_3,x_4)\in\mathbb{R}^4$. The
first two components $(x_1,x_2)$ perturb the bone position and
thickness as in \eqref{eq:example_domain_perturb}. The final two
components represent beam perturbations:
\begin{itemize}
    \item Angular deviation $x_3 \sim \mathrm{N}(0,\pi/60)$, shifting
    the central beam direction from $\theta=\pi$.
    \item Energy shift $x_4 \sim \mathrm{N}(0,5)$\,MeV, added to the
    nominal mean energy of 150\,MeV.
\end{itemize}

\begin{figure}[htbp]
    \centering
    \includegraphics[width=0.75\linewidth]{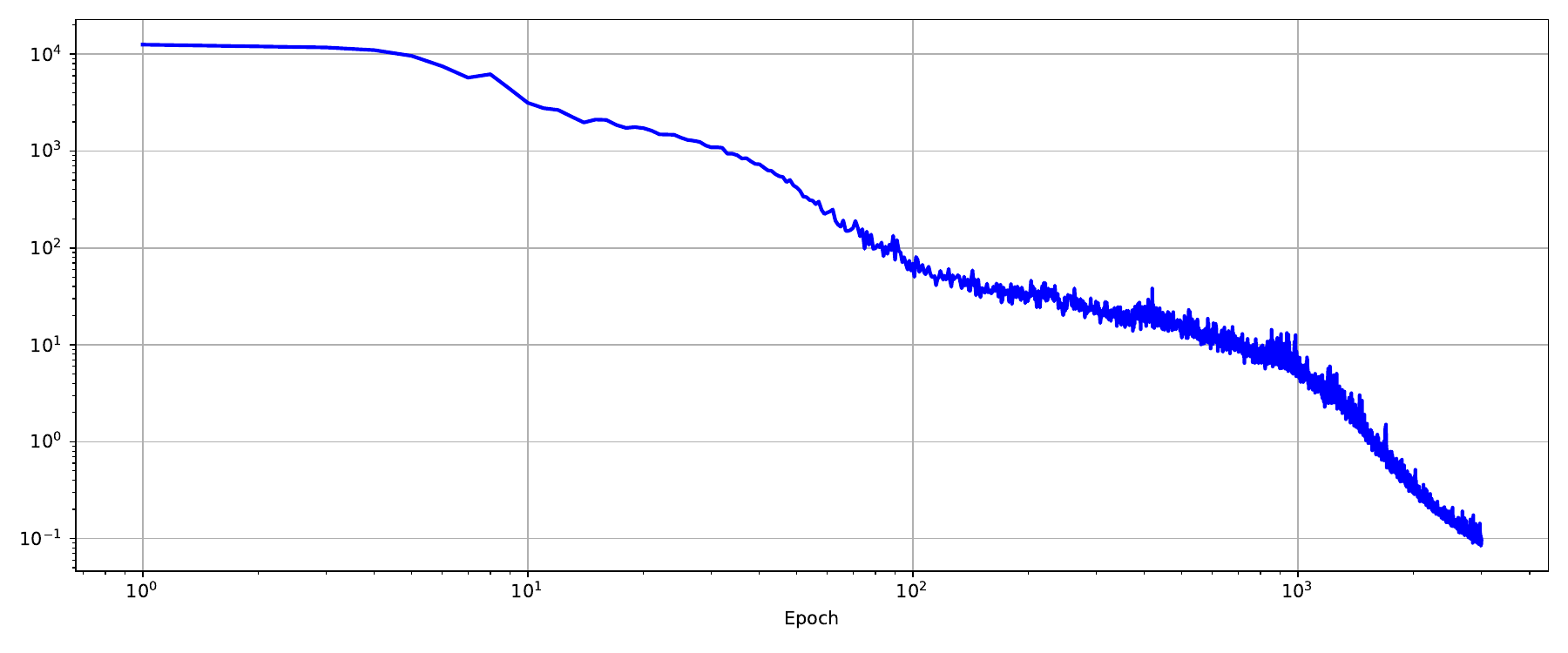}
    \caption{\textit{(Example 6)} Training history of the 4D surrogate.
    The $\ell^2$ loss between surrogate predictions $\mathscr{D}_\theta$
    and reference log-dose $\vec d$ decreases steadily, confirming
    convergence.}
    \label{fig:Experiment_6_loss}
\end{figure}

Dose is simulated in TOPAS/Geant4 with $2.5\times 10^{5}$ particle
histories, voxel grid $M_1=1500$, $M_2=200$, and the same pencil-beam
profile as Example~5. We generate $N=100$ phantoms. The surrogate has
$L_h=3$ hidden and $L_d=3$ dropout layers, width
$N_{\mathrm{width}}=512$, dropout probability
$p_{\mathrm{drop}}=0.05$, and learning rate $\eta=10^{-3}$. Figure
\ref{fig:Experiment_6_loss} shows the training loss.

\begin{figure}[htbp]
    \centering
    \includegraphics[width=0.85\linewidth]{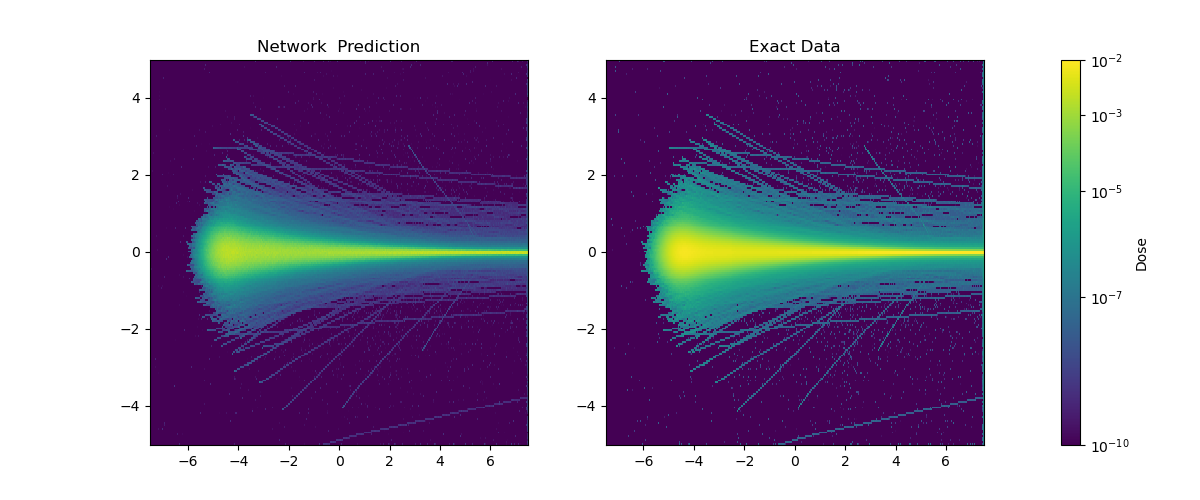}
    \caption{\textit{(Example 6)} Expected log-dose from the surrogate
    $\mathbb{E}[\mathscr{D}_\theta]$ (left) compared with Monte Carlo
    $\vec d$ (right) for $\vec x=(0,0,0,0)$. The surrogate captures the
    overall shape but underestimates the central peak.}
    \label{fig:Experiment_6_mean}
\end{figure}

\begin{figure}[htbp]
\centering
\begin{subfigure}{0.40\textwidth}
    \includegraphics[width=\textwidth]{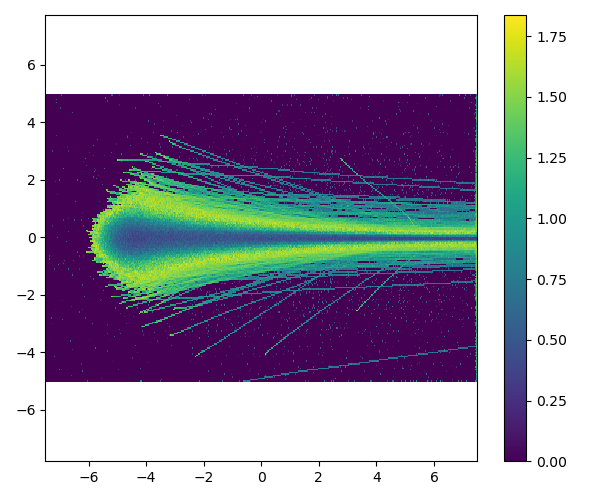}
    \caption{Parametric variance.}
    \label{fig:Experiment_6_var_param}
\end{subfigure}
\begin{subfigure}{0.40\textwidth}
    \includegraphics[width=\textwidth]{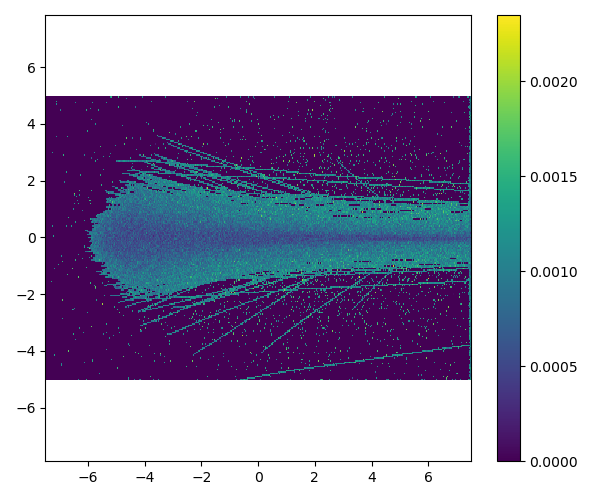}
    \caption{Epistemic variance.}
    \label{fig:Experiment_6_var_epi}
\end{subfigure}
\caption{\textit{(Example 6)} Variance decomposition for
$\vec x=(0,0,0,0)$. Parametric variance dominates and concentrates near
the distal edge and Bragg peak, while epistemic variance remains
smaller and localised.}
\label{fig:Experiment_6_variance}
\end{figure}

\begin{figure}[htbp]
    \centering
    \includegraphics[width=0.75\linewidth]{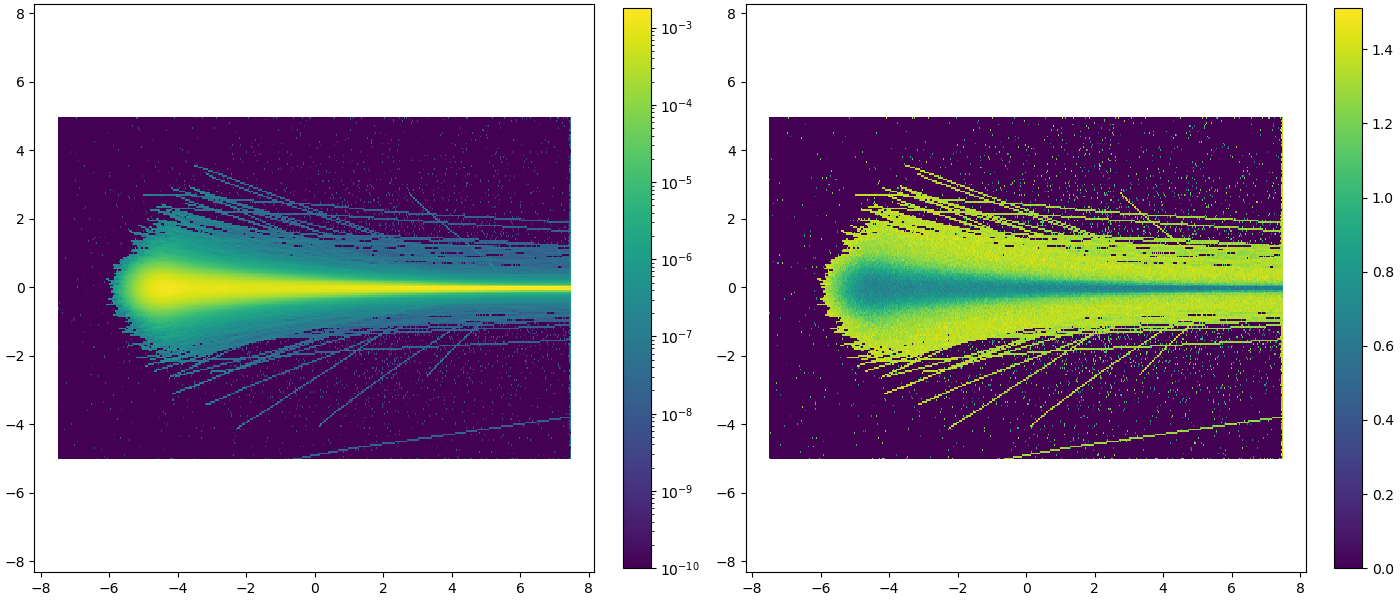}
    \caption{\textit{(Example 6)} Error maps for $\vec x=(0,0,0,0)$. Left: $|10^{\vec d(\vec x)} - 10^{\mathscr{D}(\vec x)}|$
    absolute error between surrogate and dose. Right: $|\vec d(\vec x) - \mathscr{D}(\vec x)|$ logarithmic error
    highlighting discontinuities at the edges of the beam. Errors
    concentrate in regions of high uncertainty.}
    \label{fig:Experiment_6_error}
\end{figure}

For the test input $\vec x=(0,0,0,0)$, the surrogate mean prediction
matches the Monte Carlo shape but underestimates central magnitude
(Figure~\ref{fig:Experiment_6_mean}). Variance maps
(Figure~\ref{fig:Experiment_6_variance}) show higher uncertainty before
the bone–water boundary compared with Example~5, reflecting sensitivity
to angular perturbations. Parametric variance dominates epistemic
variance, consistent with beam and geometry perturbations being the
main source of variability. Error maps
(Figure~\ref{fig:Experiment_6_error}) confirm that discrepancies
concentrate near high-uncertainty regions.

\
\subsection*{Example 7: Three-dimensional water phantom with beam uncertainty}

Finally, we test scalability to full volumetric dose prediction. The
surrogate is trained to map $\vec x\in\mathbb{R}^2$ to a
three-dimensional log-dose distribution in a homogeneous water
phantom,
\[
(-20,20)\times(-20,20)\times(-20,20)~\mathrm{cm}^3,
\]
voxelised into $M_1=M_2=M_3=60$ bins. As before, we take
$\log_{10}(\mathrm{dose}+10^{-10})$ for stability, yielding tensors
$\vec d^{(i)}\in[-10,\infty)^{M_1\times M_2\times M_3}$ from TOPAS.

The input vector $\vec x=(x_1,x_2)$ represents horizontal and vertical
shifts of the beam, mimicking patient misalignment. The beam enters at
$z=20$ with Gaussian spatial profile centred at $(x_1,y_2)$ and width
0.65\,cm. We model $x_1,x_2\sim\mathrm{N}(0,1)$\,cm. Angular spread is
fixed Gaussian with mean $\theta=0$ and width 0.0032\,rad; energy
distribution is Gaussian with mean 200\,MeV, width 3\,MeV. A total of
$10^6$ particle histories were tracked and $N=100$ phantoms simulated.

\begin{figure}[htbp]
    \centering
    \includegraphics[width=0.75\linewidth]{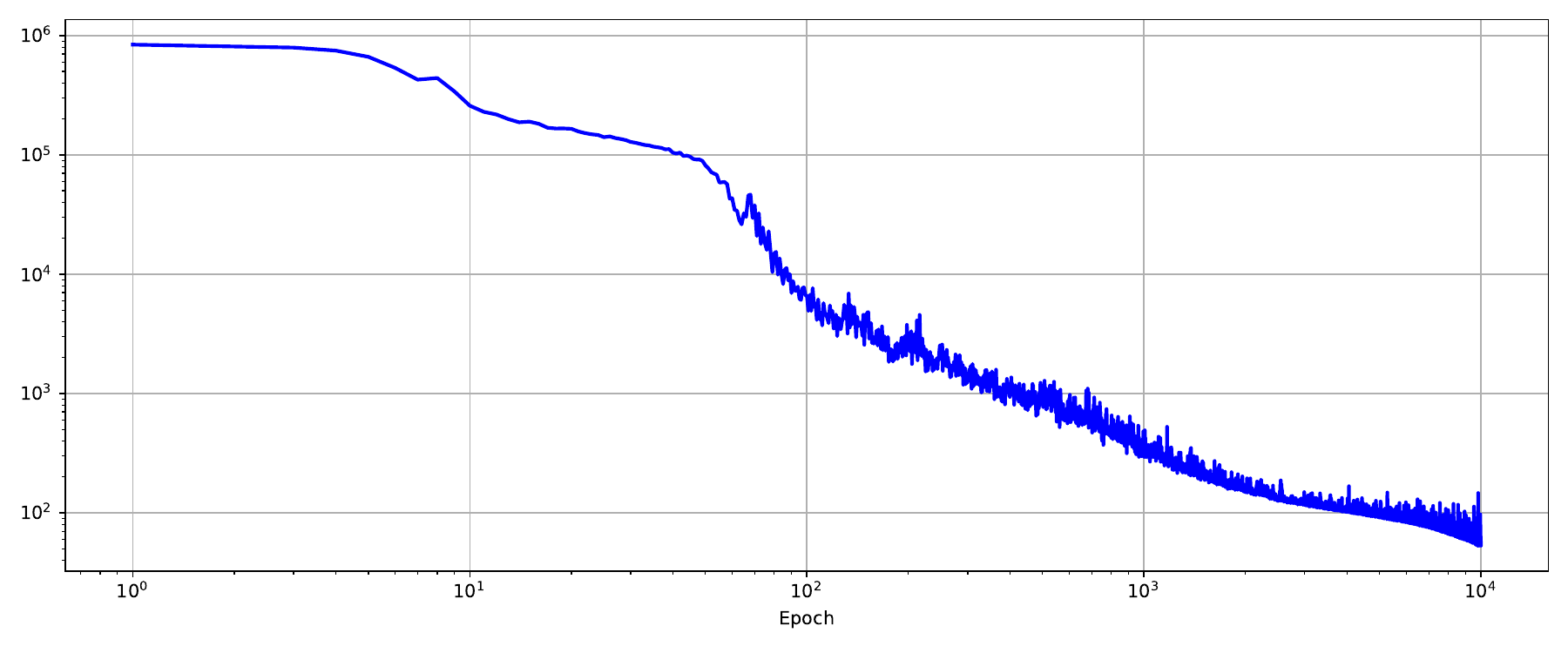}
    \caption{\textit{(Example 7)} Training history of the 3D surrogate.
    The $\ell^2$ loss between $\mathscr{D}_\theta$ and Monte Carlo
    log-dose $\vec d$ decreases steadily, confirming convergence.}
    \label{fig:Experiment_7_loss}
\end{figure}

The surrogate architecture follows earlier experiments:
$L_h=L_d=3$, hidden width $512$, dropout probability $0.05$, learning
rate $10^{-3}$. Figure~\ref{fig:Experiment_7_loss} shows the loss
history. Using an NVIDIA GeForce RTX 4090 processor we are able to estimate that the average time to complete a topas simulation was approximately $344.2$ seconds for a single instance of $\vec x$; whereas evaluation cost for the trained neural network (using dropout) was estimated to be $2.6735\times 10^{-2}$ seconds for a single instance of $\vec x$, which $\times 12000$ increase. The loading time for the neutral network model was estimated to be $2.273$ seconds.

\begin{figure}[htbp]
\centering
\begin{subfigure}{0.40\textwidth}
    \includegraphics[width=\textwidth]{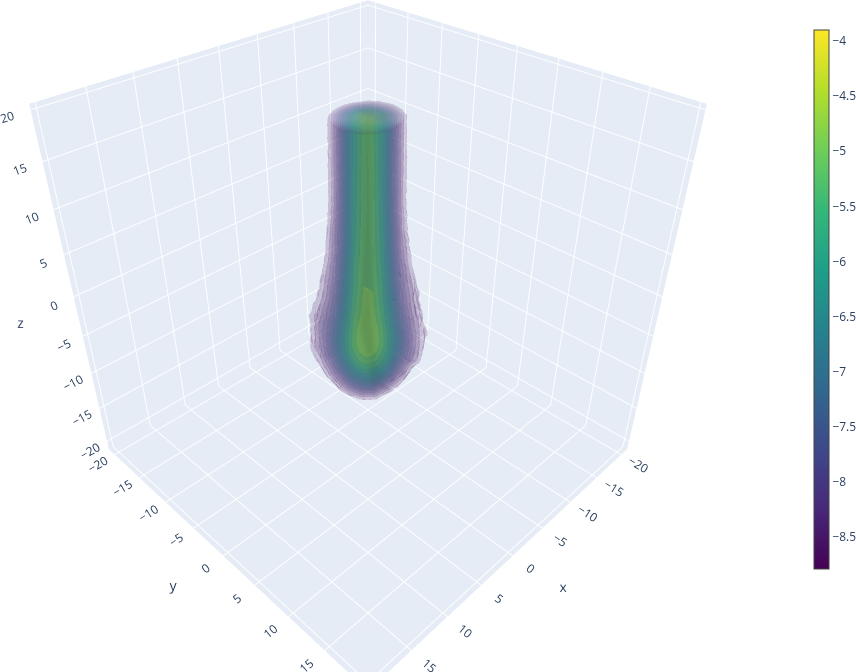}
    \caption{Expected dose $\mathbb{E}[\mathscr{D}_\theta]$.}
    \label{fig:Experiment_7_mean}
\end{subfigure}
\begin{subfigure}{0.40\textwidth}
    \includegraphics[width=\textwidth]{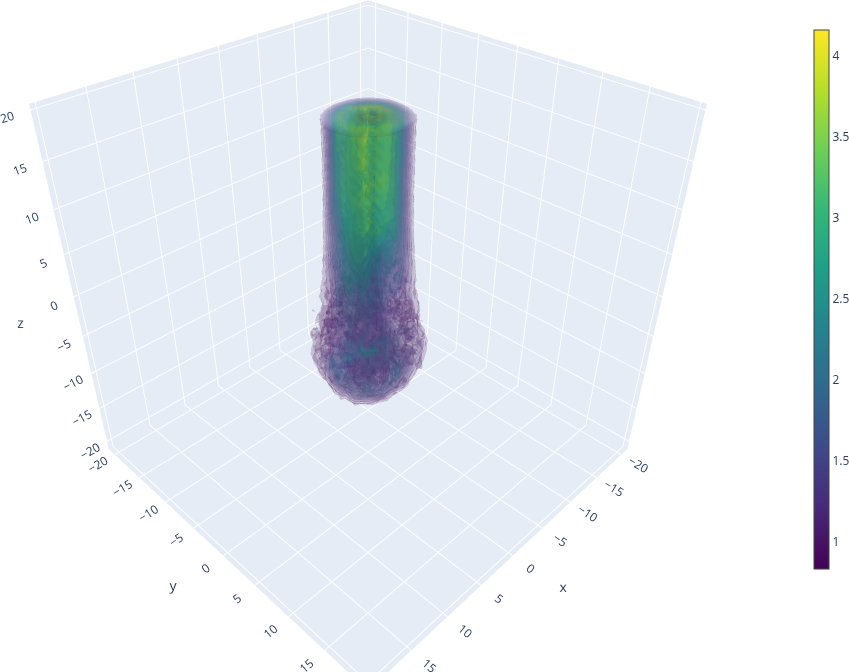}
    \caption{Dose variance $\mathrm{Var}[\mathscr{D}_\theta]$.}
    \label{fig:Experiment_7_var}
\end{subfigure}
\caption{\textit{(Example 7)} Mean and variance for $\vec x=(0,0)$.
Uncertainty concentrates in the proximal tail, reflecting sensitivity
to beam position shifts.}
\label{fig:Experiment_7_mean_var}
\end{figure}

\begin{figure}[htbp]
\centering
\begin{subfigure}{0.40\textwidth}
    \includegraphics[width=\textwidth]{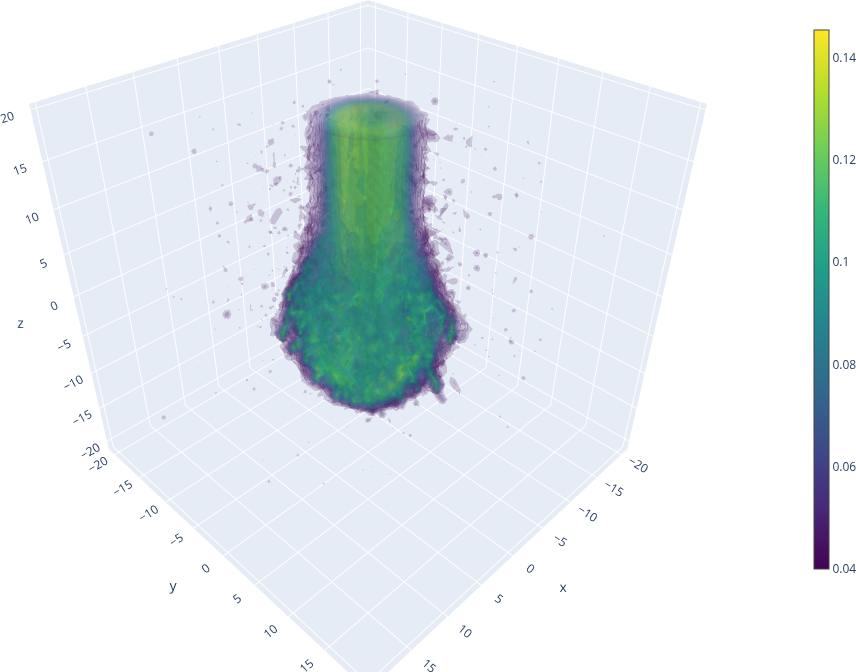}
    \caption{Parametric error.}
    \label{fig:Experiment_7_error_param}
\end{subfigure}
\begin{subfigure}{0.40\textwidth}
    \includegraphics[width=\textwidth]{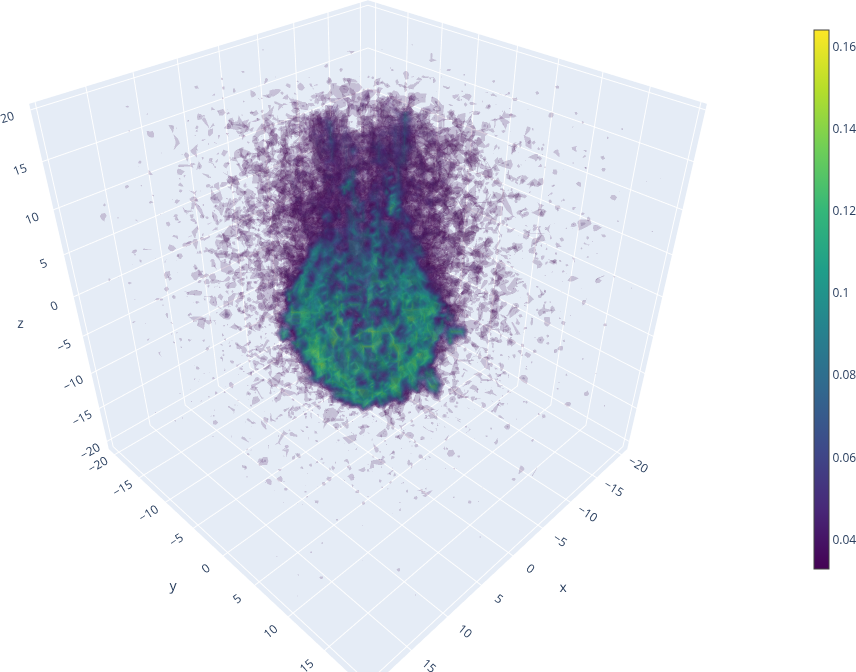}
    \caption{Epistemic error.}
    \label{fig:Experiment_7_error_epi}
\end{subfigure}
\caption{\textit{(Example 7)} Decomposition of pointwise errors for
$\vec x=(0,0)$. Parametric error dominates in the proximal tail,
whereas epistemic error concentrates near the distal Bragg surface.}
\label{fig:Experiment_7_error}
\end{figure}

For $\vec x=(0,0)$, the surrogate mean reproduces the volumetric beam
shape, while variance localises in the proximal tail
(Figure~\ref{fig:Experiment_7_mean_var}). Decomposition
(Figure~\ref{fig:Experiment_7_error}) shows parametric error dominates
in the proximal region, while epistemic error is more pronounced near
the distal fall-off and Bragg surface. This aligns with expectations:
beam misalignment drives input variability, whereas limited training
data control model uncertainty.

\section{Discussion}
\label{sec:discussion}

The numerical experiments provide evidence that the surrogate delivers
both accurate mean dose predictions and meaningful uncertainty
estimates. From a mathematical perspective, the variance-decomposition
framework clarifies when epistemic or parametric components
dominate. In the one-dimensional benchmarks (Examples~1-3), epistemic
variance captured by dropout is largest near the Bragg peak when the
training distribution is sparse, and decreases as sample size
grows. In contrast, parametric variance dominates when input
distributions are broad, as in the two-dimensional bone-water phantom
(Example~5) where domain perturbations shift the distal edge. This
confirms that the law of total variance decomposition aligns with
intuitive sources of error: epistemic variance reflects limited model
knowledge, while parametric variance reflects true variability in
patient or beam parameters.

In terms of computational trade-offs, the convergence studies
(Examples~2-4) show that relatively few training phantoms and dropout
passes are needed to stabilise the predictive mean, while additional
resources primarily reduce noise in the uncertainty estimates. This
suggests that training cost can be balanced against the desired
precision of the uncertainty maps. Moreover, calibration by
split-conformal methods further improves coverage without requiring
extra forward evaluations. Compared to full Monte Carlo, the surrogate
achieves orders-of-magnitude speedups, making it feasible for inner
optimisation loops or large scenario sets where repeated MC would be
infeasible.

Clinically, the localisation of uncertainty is highly relevant. In the
two- and three-dimensional phantoms (Examples~5-7), both total and
epistemic variance inflate at material boundaries and along the distal
fall-off. These are precisely the regions where small changes in
composition or density have the greatest effect on range, and where
clinical margins are typically introduced. The surrogate therefore
highlights regions where plan robustness is most critical, and where
clinicians may wish to prioritise full MC verification.

Finally, behaviour under distribution shift is consistent with
expectations. In the one-dimensional experiments (Example~2),
variance-inflation factors $\kappa_R$ exceed unity when test
distributions are displaced from the training mean, showing that
epistemic uncertainty correctly inflates out of distribution. In the
higher-dimensional phantoms (Examples~6-7), epistemic variance maps
also increase when beam angle or energy perturbations differ from the
training distribution. This provides a practical signal that the
surrogate is operating outside its domain of validity, an essential
property for safe deployment in planning and adaptive workflows.

Taken together, these results demonstrate that the surrogate combines
speed with principled uncertainty quantification. Mathematically it
faithfully implements variance decomposition and calibration;
computationally it delivers tractable evaluations at scale; and
clinically it highlights exactly those regions where robustness is
most critical. This positions the approach as a practical and
uncertainty-aware alternative to direct Monte Carlo in modern proton
therapy workflows.

\section{Outlook and Conclusion}
\label{sec:outlook}

We have developed a neural surrogate for proton dose calculation that
integrates Monte Carlo dropout to deliver calibrated predictive
uncertainty. Across a staged series of experiments, from analytic
one-dimensional benchmarks (Examples~1-4) to two- and
three-dimensional Monte Carlo phantoms (Examples~5-7), the surrogate
achieved accurate mean dose prediction while exposing voxelwise
uncertainty. Variance decomposition into epistemic and parametric
components, together with post-hoc conformal calibration, produced
uncertainty estimates that align with empirical coverage and inflate
appropriately under distribution shift. Importantly, uncertainty maps
localised at the distal fall-off and at material interfaces,
highlighting precisely the regions of greatest clinical sensitivity.

Several limitations should be acknowledged. All higher-dimensional
tests were conducted on simplified phantoms rather than patient CTs,
and the number of training phantoms was deliberately modest. These
choices established proof of concept but do not capture the diversity
of clinical geometries. In addition, we presented a single surrogate
architecture (although tested many). Deeper or convolutional models
may improve accuracy and calibration. Finally, dropout provides a
convenient but approximate uncertainty mechanism, and alternatives
such as ensembles or variational Bayesian methods warrant exploration.

Looking forward, three directions are natural. First, extending the
pipeline to patient CTs will test robustness in anatomically realistic
settings. Second, incorporating alternative Bayesian surrogates or
hybrid methods could strengthen calibration and expressivity. Third,
integration into robust optimisation frameworks and adaptive workflows
would enable uncertainty-aware planning and near-real-time dose
updates. Together these steps move towards a clinically deployable
surrogate that combines the speed of deep learning with the
trustworthiness required for safe proton therapy.

\section*{acknowledgements}

AP and TP are supported by the EPSRC programme grant Mathematics of
Radiation Transport (MaThRad) EP/W026899/2. TP is also grateful for
support of the Leverhulme Trust RPG-2021-238. 

\printbibliography
\end{document}

%% file: nuclear.tex
\newcommand{\proton}[1]{%
    \shade[ball color=red!80!white, draw=black, line width=0.5pt] (#1) circle (.25);
    \draw[black] (#1) node{$+$};
}

\newcommand{\neutron}[1]{%
    \shade[ball color=lime!70!green, draw=black, line width=0.5pt] (#1) circle (.25);
}

\newcommand{\electron}[3]{%
    \draw[rotate = #3, color=gray!60!white, line width=0.8pt] (0,0) ellipse (#1 and #2);
    \shade[ball color=yellow!90!white, draw=black, line width=0.5pt] (0,#2)[rotate=#3] circle (.1);
}

\newcommand{\legendelectron}[1]{%
    \shade[ball color=yellow!90!white, draw=black, line width=0.5pt] (#1) circle (.1);
}

\newcommand{\nucleus}{%
    \neutron{0.1,0.3}
    \proton{0,0}
    \neutron{0.3,0.2}
    \proton{-0.2,0.1}
    \neutron{-0.1,0.3}
    \proton{0.2,-0.15}
    \neutron{-0.05,-0.12}
    \proton{0.17,0.21}
}

\newcommand{\inelastic}[2]{
  \proton{#1,#2};
  \draw[->,thick,cyan!80!white](#1+0.5,#2)--(4,-3.7); 
  \draw[->,thick,cyan!80!white](0,-3)--(-0.3,-4); 
  \shade[ball color=yellow] (-0.3,-4) circle (.1); 
}

\newcommand{\elastic}[2]{
  \proton{#1,#2};
  \draw[->,thick,orange,bend right=90](#1+0.5,#2) to  [out=-30, in=-150] (4,3.);
}

\newcommand{\protoncollision}[3]{
  \proton{#1,#2};
  \draw[->,thick,red](#1+0.5,#2)--(-0.5,0);%
  \draw[snake=coil, line after snake=0pt, segment aspect=0,%
    segment length=5pt,color=red!80!blue] (0,0)-- +(4,2)%
  node[fill=white!70!yellow,draw=red!50!white, below=.01cm,pos=1.]%
  {$\gamma$};%
  \draw[->,thick,red](#1+0.5,#2)--(-0.5,0);%
  \draw[->,thick,red](0.5,0)--(3.7,-1.8);%
  \neutron{4,-2};  
}

\begin{tikzpicture}[scale=0.5]
    \nucleus
    \electron{1.5}{0.75}{80}
    \electron{1.2}{1.4}{260}
    \electron{4}{2}{30}
    \electron{4}{3}{180}
    \protoncollision{-6.}{0.}{160}
    \inelastic{-6.}{-2.}
    \elastic{-6.}{2.}

      \begin{scope}[shift={(8,-1)}]        
        \draw [thick,rounded corners=2pt] (0,-4) rectangle (8,4); 
        \node at (4, 3.5) {\textbf{Legend}};        
        \proton{0.5, 3}
        \node[anchor=west, font=\footnotesize] at (1.5,3) {Proton}; 
        \neutron{0.5, 2}
        \node[anchor=west, font=\footnotesize] at (1.5,2) {Neutron};         
        \legendelectron{0.5, 1}
        \node[anchor=west, font=\footnotesize] at (1.5,1) {Electron};        
        \draw[->,thick,red] (0.5,0) -- +(1,0);
        \node[anchor=west, font=\footnotesize] at (1.5,0) {Nonelastic collision};        
        \draw[->,thick,cyan!80!white] (0.5,-1) -- +(1,0);
        \node[anchor=west, font=\footnotesize] at (1.5,-1) {Inelastic interaction};         
        \draw[->,thick,orange] (0.5,-2) -- +(1,0);
        \node[anchor=west, font=\footnotesize] at (1.5,-2) {Elastic interaction}; 
        \draw[snake=coil, line after snake=0pt, segment aspect=0,
          segment length=5pt,color=red!80!blue] (0.5,-3) -- +(1,0);
        \node[anchor=west, font=\footnotesize] at (1.5,-3) {prompt-$\gamma$ emission};
      \end{scope}
\end{tikzpicture}

%% file: pipeline.tex
\begin{tikzpicture}[
  font=\small,
  block/.style={draw, rounded corners, very thick, minimum width=3.6cm, minimum height=1.25cm, align=center, fill=white},
  thinblock/.style={draw, rounded corners, thick, minimum width=3.6cm, minimum height=1.0cm, align=center, fill=white},
  data/.style={draw, dashed, rounded corners, thick, minimum width=3.6cm, minimum height=1.0cm, align=center, fill=white},
  arrow/.style={-{Latex[length=3mm,width=2mm]}, very thick},
  darr/.style={-{Latex[length=2.5mm,width=2mm]}, thick, dashed},
  lab/.style={inner sep=1pt, above}
]

\node[block] (inputs) {Inputs\\[1pt]\footnotesize Phantom, beam parameters};
\node[block, right=1.5cm of inputs] (surrogate) {Neural surrogate\\[1pt]\footnotesize with dropout};
\node[thinblock, right=1.5cm of surrogate] (ensemble) {Inference ensemble\\[1pt]\footnotesize stochastic passes};
\node[thinblock, below=1.5cm of ensemble] (reduce) {Predictive mean \\ \& variance};
\node[thinblock, right=1.5cm of reduce] (calib) {Calibration\\[1pt]\footnotesize align nominal $\leftrightarrow$ empirical};
\node[block, above=1.5cm of calib] (outputs) {Outputs\\[1pt]\footnotesize Calibrated dose \& uncertainty maps};

\node[data, below=1.5cm of inputs] (mcdata) {High‐fidelity MC dose};
\node[thinblock, right=1.65cm of mcdata] (train) {Supervised training\\[1pt]\footnotesize minimise loss};

\draw[arrow] (inputs) -- (surrogate);
\draw[arrow] (surrogate) -- (ensemble);
\draw[arrow] (ensemble) -- (reduce);
\draw[arrow] (reduce) -- (calib);
\draw[arrow] (calib) -- (outputs);

\draw[darr] (mcdata) -- (train);
\draw[darr] (train) -- (surrogate);

\node[lab] at ($(inputs.north)!0.5!(surrogate.north)$) {};
\node[lab] at ($(ensemble.north)$) {};

\node[align=left, anchor=north west] at ($(inputs.south west)+(1.5,-3.0)$) {%
\begin{tabular}{@{}ll@{\hspace{2em}}ll@{\hspace{2em}}ll@{\hspace{2em}}ll@{}}
\raisebox{0.6ex}{\tikz{\node[block, minimum width=0.5cm, minimum height=0.3cm]{};}} 
  & Processing component &
\raisebox{0.6ex}{\tikz{\node[data, minimum width=0.5cm, minimum height=0.3cm]{};}} 
  & Data source (MC) &
\raisebox{0.6ex}{\tikz{\draw[arrow] (0,0)--(0.8,0);}} 
  & Inference flow &
\raisebox{0.6ex}{\tikz{\draw[darr] (0,0)--(0.8,0);}} 
  & Training/supervision
\end{tabular}

};

\end{tikzpicture}

%% file: nn.tex
\begin{tikzpicture}[
    neuron/.style={circle, draw, minimum size=8mm, fill=blue!10},
    dropout/.style={circle, draw, dashed, minimum size=8mm, fill=red!10},
    layer/.style={rectangle, draw=none, minimum height=1cm},
    arrow/.style={-{Latex}, thick},
    every label/.append style={font=\footnotesize}
]

\node[neuron, label=left:Input] (x) at (0,0) {$\vec x$};

\node[neuron] (h0) at (2,2) {};
\node[neuron] (h1) at (2,1) {};
\node[neuron] (h2) at (2,0) {};
\node[neuron] (h3) at (2,-1) {};
\node[neuron] (h4) at (2,-2) {};

\node[neuron] (h5) at (4,2) {};
\node[neuron] (h6) at (4,1) {};
\node[neuron] (h7) at (4,0) {};
\node[neuron] (h8) at (4,-1) {};
\node[neuron] (h9) at (4,-2) {};

\node[neuron] (d0) at (6,2) {};
\node[neuron] (d1) at (6,1) {};
\node[neuron] (d2) at (6,0) {};
\node[neuron] (d3) at (6,-1) {};
\node[neuron] (d4) at (6,-2) {};
\node[neuron] (d5) at (8,2) {};
\node[neuron] (d6) at (8,1) {};
\node[neuron] (d7) at (8,0) {};
\node[neuron] (d8) at (8,-1) {};
\node[neuron] (d9) at (8,-2) {};
\node[neuron, label=center:$\mathscr{D}_{\theta}$] (y) at (10,0) {};

\draw[arrow] (x) -- (h0);
\draw[arrow] (x) -- (h1);
\draw[arrow] (x) -- (h2);
\draw[arrow] (x) -- (h3);
\draw[arrow] (x) -- (h4);
\draw[arrow] (h0) -- (h5);
\draw[arrow] (h0) -- (h6);
\draw[arrow] (h0) -- (h7);
\draw[arrow] (h0) -- (h8);
\draw[arrow] (h0) -- (h9);
\draw[arrow] (h1) -- (h5);
\draw[arrow] (h1) -- (h6);
\draw[arrow] (h1) -- (h7);
\draw[arrow] (h1) -- (h8);
\draw[arrow] (h1) -- (h9);
\draw[arrow] (h2) -- (h5);
\draw[arrow] (h2) -- (h6);
\draw[arrow] (h2) -- (h7);
\draw[arrow] (h2) -- (h8);
\draw[arrow] (h2) -- (h9);
\draw[arrow] (h3) -- (h5);
\draw[arrow] (h3) -- (h6);
\draw[arrow] (h3) -- (h7);
\draw[arrow] (h3) -- (h8);
\draw[arrow] (h3) -- (h9);
\draw[arrow] (h4) -- (h5);
\draw[arrow] (h4) -- (h6);
\draw[arrow] (h4) -- (h7);
\draw[arrow] (h4) -- (h8);
\draw[arrow] (h4) -- (h9);
\draw[arrow] (h5) -- (d0);
\draw[arrow] (h5) -- (d1);
\draw[arrow] (h5) -- (d2);
\draw[arrow] (h5) -- (d3);
\draw[arrow] (h5) -- (d4);
\draw[arrow] (h6) -- (d0);
\draw[arrow] (h6) -- (d1);
\draw[arrow] (h6) -- (d2);
\draw[arrow] (h6) -- (d3);
\draw[arrow] (h6) -- (d4);
\draw[arrow] (h7) -- (d0);
\draw[arrow] (h7) -- (d1);
\draw[arrow] (h7) -- (d2);
\draw[arrow] (h7) -- (d3);
\draw[arrow] (h7) -- (d4);
\draw[arrow] (h8) -- (d0);
\draw[arrow] (h8) -- (d1);
\draw[arrow] (h8) -- (d2);
\draw[arrow] (h8) -- (d3);
\draw[arrow] (h8) -- (d4);
\draw[arrow] (h9) -- (d0);
\draw[arrow] (h9) -- (d1);
\draw[arrow] (h9) -- (d2);
\draw[arrow] (h9) -- (d3);
\draw[arrow] (h9) -- (d4);
\draw[arrow] (d0) -- (d5);
\draw[arrow] (d0) -- (d6);
\draw[arrow] (d0) -- (d7);
\draw[arrow] (d0) -- (d8);
\draw[arrow] (d0) -- (d9);
\draw[arrow] (d1) -- (d5);
\draw[arrow] (d1) -- (d6);
\draw[arrow] (d1) -- (d7);
\draw[arrow] (d1) -- (d8);
\draw[arrow] (d1) -- (d9);
\draw[arrow] (d2) -- (d5);
\draw[arrow] (d2) -- (d6);
\draw[arrow] (d2) -- (d7);
\draw[arrow] (d2) -- (d8);
\draw[arrow] (d2) -- (d9);
\draw[arrow] (d3) -- (d5);
\draw[arrow] (d3) -- (d6);
\draw[arrow] (d3) -- (d7);
\draw[arrow] (d3) -- (d8);
\draw[arrow] (d3) -- (d9);
\draw[arrow] (d4) -- (d5);
\draw[arrow] (d4) -- (d6);
\draw[arrow] (d4) -- (d7);
\draw[arrow] (d4) -- (d8);
\draw[arrow] (d4) -- (d9);
\draw[arrow] (d5) -- (y);
\draw[arrow] (d6) -- (y);
\draw[arrow] (d7) -- (y);
\draw[arrow] (d8) -- (y);
\draw[arrow] (d9) -- (y);

\end{tikzpicture}

%% file: nndp.tex
\begin{tikzpicture}[
    neuron/.style={circle, draw, minimum size=8mm, fill=blue!10},
    dropout/.style={circle, draw, dashed, minimum size=8mm, fill=red!10},
    layer/.style={rectangle, draw=none, minimum height=1cm},
    arrow/.style={-{Latex}, thick},
    every label/.append style={font=\footnotesize}
]

\node[neuron, label=above:Input] (x) at (0,0) {$x$};

\node[dropout] (h0) at (2,2) {};
\node[dropout] (h1) at (2,1) {};
\node[neuron] (h2) at (2,0) {};
\node[neuron] (h3) at (2,-1) {};
\node[dropout] (h4) at (2,-2) {};

\node[neuron] (h5) at (4,2) {};
\node[dropout] (h6) at (4,1) {};
\node[dropout] (h7) at (4,0) {};
\node[neuron] (h8) at (4,-1) {};
\node[dropout] (h9) at (4,-2) {};

\node[dropout] (d0) at (6,2) {};
\node[neuron] (d1) at (6,1) {};
\node[dropout] (d2) at (6,0) {};
\node[dropout] (d3) at (6,-1) {};
\node[neuron] (d4) at (6,-2) {};
\node[neuron, label=above:Hidden layer] (d5) at (8,2) {};
\node[neuron] (d6) at (8,1) {};
\node[neuron] (d7) at (8,0) {};
\node[neuron] (d8) at (8,-1) {};
\node[neuron] (d9) at (8,-2) {};
\node[neuron, label=above:Output] (y) at (10,0) {$\mathscr{D}_{\theta^{(t)}}$};

\draw[arrow] (x) -- (h2);
\draw[arrow] (x) -- (h3);
\draw[arrow] (h2) -- (h5);
\draw[arrow] (h2) -- (h8);
\draw[arrow] (h3) -- (h5);
\draw[arrow] (h3) -- (h8);
\draw[arrow] (h5) -- (d1);
\draw[arrow] (h5) -- (d4);
\draw[arrow] (h8) -- (d1);
\draw[arrow] (h8) -- (d4);
\draw[arrow] (d1) -- (d5);
\draw[arrow] (d1) -- (d6);
\draw[arrow] (d1) -- (d7);
\draw[arrow] (d1) -- (d8);
\draw[arrow] (d1) -- (d9);
\draw[arrow] (d4) -- (d5);
\draw[arrow] (d4) -- (d6);
\draw[arrow] (d4) -- (d7);
\draw[arrow] (d4) -- (d8);
\draw[arrow] (d4) -- (d9);
\draw[arrow] (d5) -- (y);
\draw[arrow] (d6) -- (y);
\draw[arrow] (d7) -- (y);
\draw[arrow] (d8) -- (y);
\draw[arrow] (d9) -- (y);

\node at ($(h5)!0.5!(d4) + (-1,2.75)$) {$\overbrace{\hspace{4.5cm}}^{\text{Dropout layers}}$};
\end{tikzpicture}